\documentclass[journal]{IEEEtran}
\usepackage{soul}
\usepackage{wrapfig}
\usepackage{amssymb}
\usepackage{graphicx}
\usepackage[export]{adjustbox}
\usepackage{graphicx}
\usepackage{lipsum}
\usepackage{amsfonts}
\usepackage{amsmath}
\usepackage{subcaption}
\usepackage[table,dvipsnames]{xcolor}
\newcolumntype{L}{@{}>{\kern\tabcolsep}l<{\kern\tabcolsep}}
\usepackage{booktabs}
\usepackage{multirow, makecell}
\usepackage{enumitem}
\usepackage[misc]{ifsym}
\usepackage{algorithm,algorithmic}
\usepackage[table]{xcolor}
\usepackage{float}  
\usepackage{bm}
\renewcommand{\figurename}{Figure}

\usepackage{url}

\usepackage{tabularx}
\usepackage{colortbl}
\usepackage{arydshln,array}

\usepackage{xpatch,letltxmacro}
\usepackage{dblfloatfix}    

\newcolumntype{P}[1]{>{\centering\arraybackslash}p{#1}}

\usepackage{url}
\urldef{\mailsa}\path|{******, *******, *******, *******,|
\urldef{\mailsb}\path|********, *******, *******, *******,|
\urldef{\mailsc}\path|********, *********, *****}*******|

\newcommand\Tstrut{\rule{0pt}{4.0ex}}       
\newcommand\Bstrut{\rule[-2ex]{0pt}{0pt}} 
\newcommand{\TBstrut}{\Tstrut\Bstrut} 

\makeatletter

\newcommand{\ALC@it@nostep}{\item[]}
\LetLtxMacro\oldalgorithmic\algorithmic
\LetLtxMacro\endoldalgorithmic\endalgorithmic
\renewenvironment{algorithmic}[1][0]{%
  \oldalgorithmic[#1]%
  \xpatchcmd{\STATE}{\ALC@it}{\ALC@it@nostep}{}{}%
  \xpatchcmd{\FOR}{\ALC@it}{\ALC@it@nostep \textbf{\textit{Step 1 -- }}}{}{}%
  \xpatchcmd{\RETURN}{\ALC@it}{\ALC@it@nostep}{}{}%
  \xpatchcmd{\ENDFOR}{\ALC@it}{\ALC@it@nostep}{}{}%

}{\endoldalgorithmic}
\makeatother

\ifCLASSINFOpdf
\else
\fi
\begin{document}
%
\title{Deep Geodesic Learning for Segmentation and Anatomical Landmarking}
\author{Neslisah~Torosdagli,
Denise K. Liberton,
Payal Verma,
Murat Sincan,
Janice S. Lee,
        and~Ulas~Bagci,~\IEEEmembership{Senior Member,~IEEE,}
\thanks{Corresponding author: ulasbagci@gmail.com}
\thanks{D. Liberton, P. Verma, and J. Lee are with the Craniofacial Anomalies and Regeneration section, National Institute of Dental and Craniofacial Research (NIDCR), National Institutes of Health (NIH), Bethesda, MD.}
\thanks{M. Sincan is with University of South Dakota Sanford School of Medicine, Sioux Falls, SD.}
\thanks{N. Torosdagli and U. Bagci are with Center for Research in Computer Vision at University of Central Florida, Orlando, FL.}
}

\markboth{IEEE Transactions on Medical Imaging}%
{Torosdagli \MakeLowercase{\textit{et al.}}}
%



\maketitle

\begin{abstract}
In this paper, we propose a novel deep learning framework for anatomy segmentation and automatic landmarking. Specifically, we focus on the challenging problem of mandible segmentation from cone-beam computed tomography (CBCT) scans and identification of 9 anatomical landmarks of the mandible on the geodesic space. The overall approach employs three inter-related steps. In step 1, we propose a deep neural network architecture with carefully designed regularization, and network hyper-parameters to perform image segmentation without the need for data augmentation and complex  post-processing refinement. In step 2, we formulate the landmark localization problem directly on the geodesic space for sparsely-spaced anatomical landmarks. In step 3, we propose to use a long short-term memory (LSTM) network to identify closely-spaced landmarks, which is rather difficult to obtain using other standard detection networks. The proposed fully automated method  showed superior efficacy compared to the state-of-the-art mandible segmentation and landmarking approaches in craniofacial anomalies and diseased states. We used a very challenging CBCT dataset of 50 patients with a high-degree of craniomaxillofacial (CMF) variability that is realistic in clinical practice. Complementary to the quantitative analysis, the qualitative visual inspection was conducted for distinct CBCT scans from 250 patients with high anatomical variability. We have also shown feasibility of the proposed work in an independent dataset from MICCAI Head-Neck Challenge (2015) achieving the state-of-the-art performance. Lastly, we present an in-depth analysis of the proposed deep networks with respect to the choice of hyper-parameters such as pooling and activation functions.
\end{abstract}


%
\IEEEpeerreviewmaketitle

\section{Introduction}
In the United States, there are more than $17$ million patients with congenital or developmental deformities of the jaws, face, and skull, also defined as the craniomaxillofacial (CMF) region~\cite{Xia2009}. Trauma, deformities from tumor ablation, and congenital birth defects are some of the leading causes of CMF deformities~\cite{Xia2009}. The number of patients who require orthodontic treatment is far beyond this number. Among CMF conditions, the mandible is one of the most frequently deformed or injured regions, with $76\%$ of facial trauma affecting the mandibular region~\cite{ARMOND2017716}.

The ultimate goal of clinicians is to provide accurate and rapid clinical interpretation, which guides appropriate treatment of CMF deformities. Cone-beam computed tomography (CBCT) is the newest conventional imaging modality for the diagnosis and treatment planning of patients with skeletal CMF deformities. Not only do CBCT scanners expose patients to lower doses of radiation compared to spiral CT scanners, but also CBCT scanners are compact, fast and less expensive, which makes them widely available. On the other hand, CBCT scans have much greater noise and artifact presence, leading to challenges in image analysis tasks.

\begin{figure}[t] 
 \centering
	\resizebox{\linewidth}{!}{
		\includegraphics[height=4.5cm]{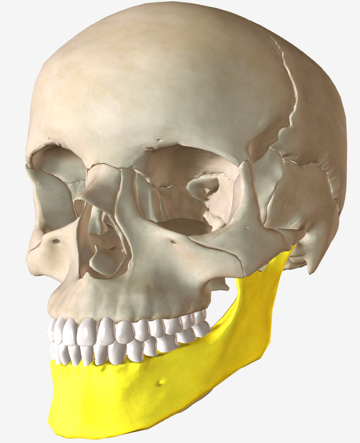}
		\quad
		\includegraphics[height=4.5cm]{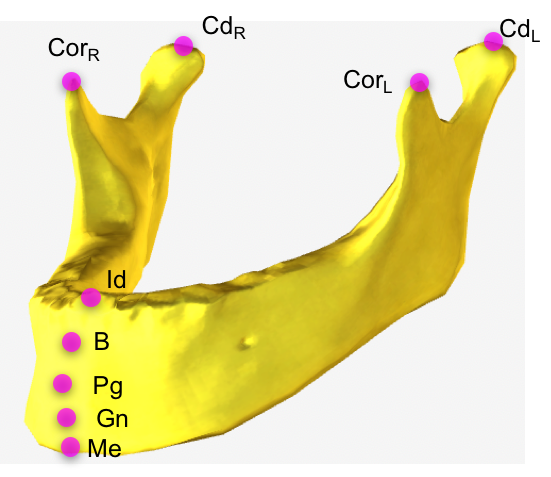}
	}
\caption{Anatomical landmarks on the mandible: Menton $(Me)$, Gnathion $(Gn)$, Pogonion $(Pg)$, B Point $(B)$, Infradentale $(Id)$, Condylar Left $(Cd_L)$, Condylar Right $(Cd_R)$, Coronoid Left $(Cor_L)$, and Coronoid Right $(Cor_R)$. We aim to locate these landmarks automatically.}
\label{fig:mandibularLandmarks}
\end{figure}

CBCT-based image analysis plays a significant role in diagnosing a disease or deformity, characterizing its severity, planning the treatment options, and estimating the risk of potential interventions. The core image analysis framework involves the detection and measurement of deformities, which requires precise segmentation of CMF bones. Landmarks, which identify anatomically distinct locations on the surface of the segmented bones, are placed and measurements are performed to determine the severity of the deformity compared to traditional 2D norms as well as to assist in treatment and surgical planning. Figure~\ref{fig:mandibularLandmarks} shows nine anatomical landmarks defined on the mandible. 

Surgical planning, patient-specific prediction of deformities, and quantification as well as clinical assessment of the deformities require precise segmentation and anatomical landmarking. However, automatically segmenting bones from the CMF regions, and accurately identifying clinically relevant anatomical landmarks on the surface of these bones continue to be a significant challenge and a persistent  problem. Currently, the landmarks have not evolved from traditional 2D anatomical landmarks for cephalometric analysis though 3D imaging has become more commonplace for clinical application.  Additionally, landmarking on CT images is tedious and manual or semi-automated and prone to operator variability. Despite some recent elaborative efforts towards making a fully automated and accurate software for segmentation of bones and landmarking for deformation analysis in dental applications \cite{zhang_2017,automated_landmarking_challanges}, the problem remains largely unsolved for global CMF deformity analysis, especially for those who have congenital or developmental deformities for whom the diagnosis and treatment planning are most critically needed.
 
\begin{figure*}[t]
\centering
\includegraphics[height=4.3cm]{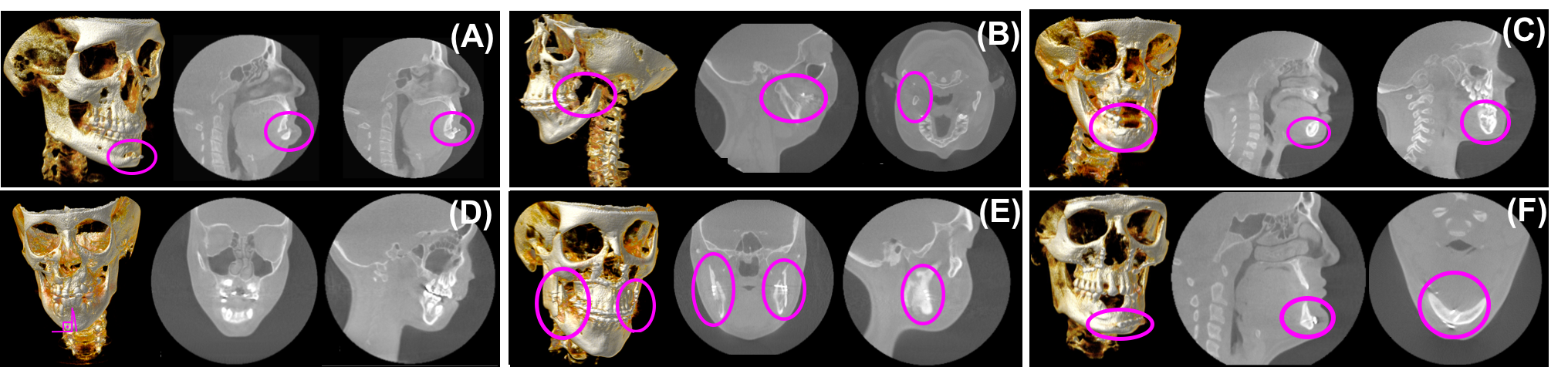}
 \caption{Examples of diverse CMF conditions are illustrated. (A) Surgical treatment, genioplasty with resultant chin advancement and fixation plate (implant) (adult), (B) missing condyle-ramus unit in the mandible in left dominant hemifacial microsomia (adult), (C) unerupted teeth in the anterior mandible with distorted anatomy (pediatric), (D) mid-sagittal plane with respect to lower jaw incisors have a serious degradation from the 90 degrees (pediatric), (E) bilateral bicortical positional screws (implants) in the ascending ramus of the mandible for rigid fixation after a bileteral sagittal split osteotomy (adult), (F) plate and screws (implants) in the anterior mandible for rigid fixation and reduction of an oblique fracture} (adult).
\label{fig:anomalies}
\end{figure*}

The main reason for this research gap is high anatomical variability in the shape of these bones due to their deformities in such patient populations. Figure~\ref{fig:anomalies} shows some of the known CMF deformities and artifacts, including missing bones (hence missing landmarks) or irregularities from the underlying disease or the surgical treatment (Figures~\ref{fig:anomalies}a-\ref{fig:anomalies}b), varying number of teeth including missing teeth and unerupted deciduous teeth distorting the anatomy (Figures~\ref{fig:anomalies}c-\ref{fig:anomalies}d), and surgical interventions such as implants or surgical plates and screws that are necessary to treat the injury or deformity (Figures~\ref{fig:anomalies}e-\ref{fig:anomalies}f). Other reasons are image/scanner based artifacts/problems such as noise, inhomogeneity, truncation, beam hardening, and low resolution. Unlike existing methods focusing on dental applications with relatively small anatomical variations, there is a strong need for creating a general purpose, automated CMF image analysis platform that can help clinicians create a segmentation model and find anatomical landmarks for extremely challenging CMF deformities. Due to the lack of a general purpose image analysis platform, clinicians still perform their analysis either manually or semi-automatically with limited software support. This process is extremely tedious and prone to reproducibility errors.

\begin{figure*}[t]
\centering
\includegraphics[width=\linewidth]{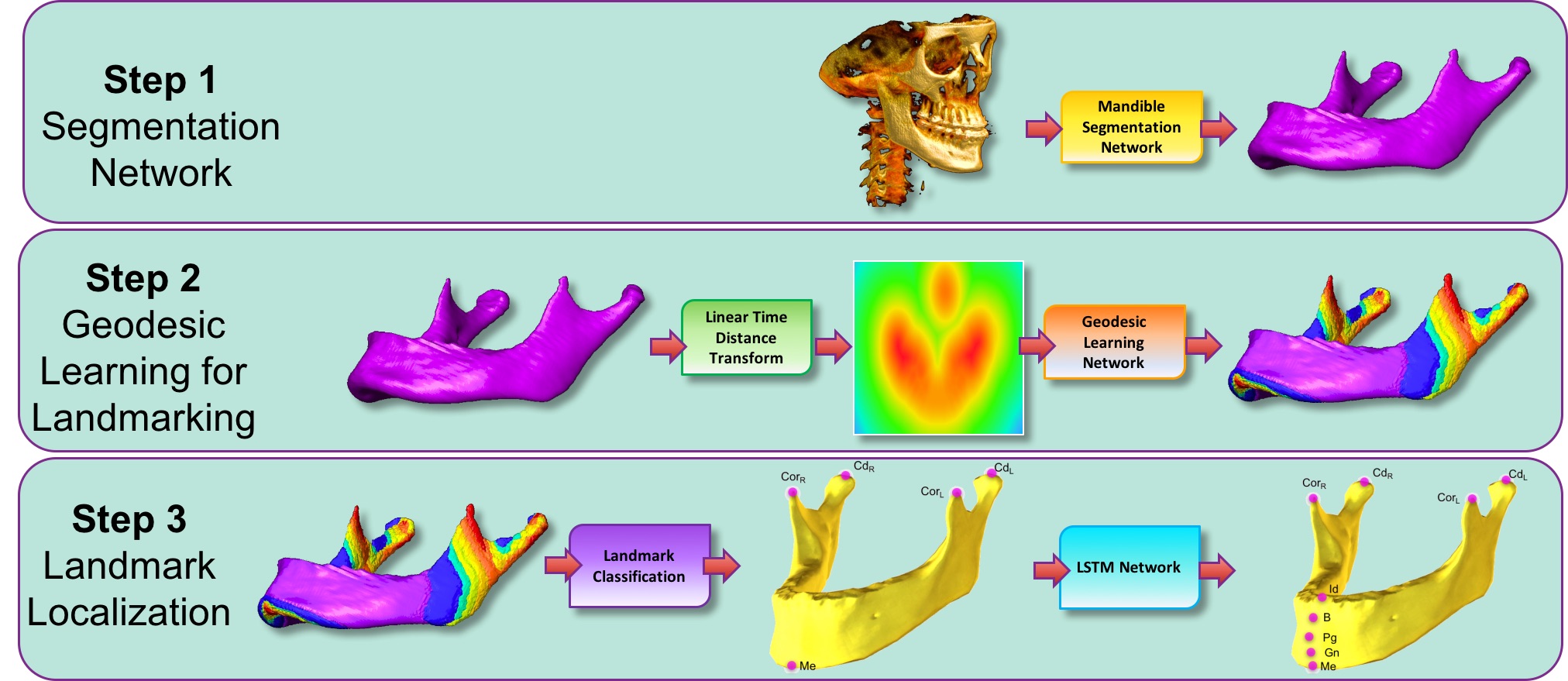}
\caption{The framework implemented in this paper starts with Fully Convolutional DenseNet for Mandible Segmentation. Following the Mandible Segmentation, Linear Time Distance Transform (LTDT) of the Mandible Bone is generated. A second U-Net~\cite{unet2015} is used for Mandibular Geodesic Learning, which transforms LTDT into combined Geodesic Map of the mandibular landmarks Menton $(Me)$, Condylar Left $(Cd_L)$, Condylar Right $(Cd_R)$, Coronoid Left $(Cd_L)$, and Coronoid Right $(Cd_R)$. After classification of $5$ Mandibular Landmarks, an LSTM Network  is used to detect Infradentale $(Id)$, B point $(B)$, Pogonion $(Pg)$, and Gnathion $(Gn)$ mandibular landmarks according to the detected position of the Menton $(Me)$ landmark. All algorithms in this proposed pipeline run in pseudo-3D (slice-by-slice $2D$). To ease understanding and visualization of the segmentation results, surface rendered volumes are presented instead of contour based binary images.}
\label{fig:pipeline}
\end{figure*}
\begin{figure*}[t]
\captionsetup{labelfont={color=black},font={color=black}}
\centering
\includegraphics[width=\linewidth]{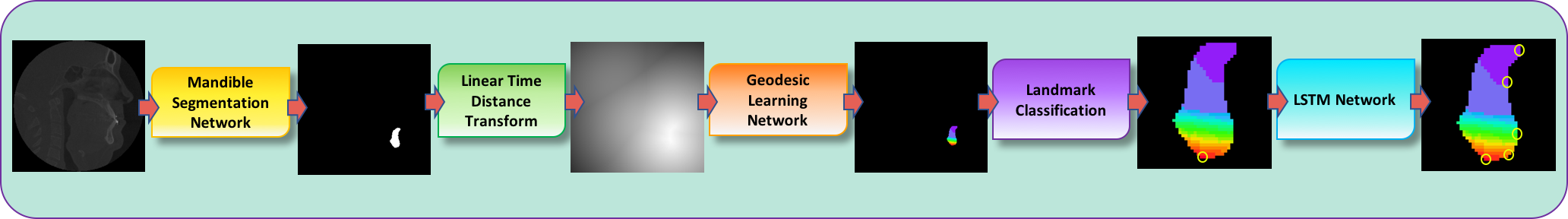}
\caption{The example workflow of a single slice in the proposed pipeline (Figure~\ref{fig:pipeline}). The outputs of the steps (Landmark Classification and LSTM Network) are zoomed in for visual illustration of the process.}
\label{fig:pipelineSingleSlice}
\end{figure*}

Our study focuses on developing a fully automated mandible segmentation and anatomical landmark localization method using CBCT scans, which is robust to challenging CMF anomalies that have the greatest need for image analysis but have the abnormal anatomy that does not fall within the normal shape parameters. Our dataset includes patients with congenital deformities fading to extreme developmental variations in CMF bones. The patient population is highly diverse, consisting of a wide range of ages across both sexes, imposing additional anatomical variability apart from the deformities. The following image-based variations have also been confirmed in our dataset: aliasing artifacts due to braces, metal alloy surgical implants (screws and plates), dental fillings, and missing bones or teeth. The overarching goal of our study is to develop a fully-automated image analysis software for mandible segmentation and anatomical landmarking that can overcome the highly variable clinical phenotypes in the CMF region.  This program will facilitate the ease of clinical application and permit the quantitative analysis that is currently tedious and prohibitive in $3D$ cephalometrics and geometric morphometrics. To this end, we include a landmarking process as a part of the segmentation algorithm to make geometric measurements more accurate, easier, and faster than manual methods. Our proposed novel deep learning algorithm includes three inter-connected steps (See Figures~\ref{fig:pipeline} and~\ref{fig:pipelineSingleSlice} for the overview of the proposed method and a sample processing pipeline for a single CBCT scan). For the first step, we design a new convolutional neural network (CNN) architecture for mandibular bone segmentation from 3D CBCT scans. For Step 2, we present a learning-based geodesic map generation algorithm for each anatomical landmark defined on the mandible. For Step 3, inspired by the success of recurrent neural networks (RNN) for capturing temporal information, we demonstrate a long short-term memory (LSTM) based algorithm modeling to capture the relationship between anatomical landmarks as a sequential learning process. 


\section{Related Work}
The mandible is the lower jaw bone and it is the only mobile bone in the CMF region. It is the largest, the strongest, and the most complex bone in the CMF region that houses the lower teeth as well as canals with blood vessels and nerves. Due to its complex structure and the significant structural variations of patients with CMF disorders, segmentation and landmark localization in the mandibular region is a very challenging problem (See Figure~\ref{fig:anomalies}). Although, there are efforts with promising performances, speeds and accuracies \cite{zhang_2017, zhang_2016, dinggang_2015,URSCHLER201823}, the literature still lacks a fully-automated, fast, and generalized software solution in response to a wide range of patient ages, deformities, and the imaging artifacts. Hence, the current convention used in clinics is either manual segmentation and annotations, or semi-automated with software support such as (in alphabetical order) 3dMDvultus (3dMD, Atlanta, Ga), Dolphin Imaging (Dolphin Imaging, Chatsworth, Ca), and InVivoDental (Anatomage, San Jose, Ca). 


Over the past decade, there have been significant improvements in mandible segmentation and landmarking using registration-based (atlas-based), model-based, and more recently machine learning-based approaches~\cite{dinggang_2017_survey}. Although, registration-based methods have been reported to achieve relatively high accuracy when shape and appearance information are integrated, these algorithms perform poorly when there are variations due to different age patients (ie. pediatrics vs. adults), missing teeth, missing parts of the region of interest, and imaging artifacts~\cite{automated_landmarking_challanges, registration_landmarking, dir_cons}. In 2015, Gupta et al.~\cite{automated_landmarking_challanges_1} developed a knowledge-based algorithm to localize 20 anatomical landmarks on the CBCT scans. Despite the promising results, the algorithm starts with the seed detection on the anterior-inferior region of the mandible and based on a template registration. In cases of missing lower incisors, mandible fractures, or other anatomical deformities that directly alter the anterior mandible, an error in seed localization can lead to a sub-optimal outcome. In 2016, Zhang et al.~\cite{zhang_2016} digitized CMF landmarks on CBCT scans using a regression forest-based landmark detector. Image segmentation was used as a guidance to address the spatial coherence of landmarks. The authors obtained a mean digitization error less than 2mm for 15 CMF landmarks. Later in 2017, Zhang et al.~\cite{zhang_2017} improved their method by proposing a joint CMF bone segmentation and landmark digitization framework via a context-guided multi-task fully convolutional neural network (FCN) adopting a U-Net architecture (i.e., the most commonly used deep network for segmentation). The spatial context of the landmarks were grasped using 3D displacement maps. An outstanding segmentation accuracy (dice similarity coefficient of $93.27\pm 0.97 \%$) was obtained along with a mean digitization error of less than 1.5 mm for identifying 15 CMF landmarks. Despite these promising performances, the study had the limitation of working on small number of landmarks due to memory constraints. That is, if there are $N_l$ landmarks and each patient's $3D$ scan is composed of $V$ voxels, 3D displacement maps use  $3\times N_l\times V$ memory as input to the $2^{nd}$ U-Net. Furthermore, most of the slices in the displacement maps were the same due to planar spatial positions, leading to inefficiency because of redundant information. 

Later, the same year, Zhang et al.~\cite{zhang_october_2017} proposed a two-stage task-oriented deep learning method for anatomical landmark detection in the brain and prostate with limited medical imaging data. The goal of the first stage was to capture the inherent associations between the local image patches and their displacements, while in the second stage the landmark locations were predicted using another CNN. Although the target region and modality of the data are different and not comparable to ours, this study is still worth mentioning due to similarity of the techniques proposed herein. The authors achieved a mean error of 2.96 mm in brain landmark localization using MR scans and a mean error of 3.34 mm in prostate landmark localization using CT scans. A potential drawback of this study is the requirement of a large amount of memory for generating the displacement maps. 






In a more conventional way, Urschler et al.~\cite{URSCHLER201823} combined image appearance information and geometric landmark configuration into a unified random forest framework, and performed an optimization procedure (the coordinate descent algorithm) that iteratively refines landmark locations jointly. The regularization strategy was also proposed to handle the combination of the appearance and the geometric landmark configuration. The authors achieved a high performance on MRI data with only a small percentage of outliers. 


\subsection{\textbf{Our Contribution}}
To date, little research has been carried out involving deep learning-based segmentation of CMF bones and landmarking. Herein, we demonstrate in-depth mandible segmentation and landmarking in a fully automated way, and we propose novel techniques that enhance accuracy and efficiency to improve the state-of-the-art approaches in the setting of high degree of anatomical variability. The latter function is highly critical as previous methods have been developed based on optimized and normal patient cohorts, but the limitations of these methods are evident in diseased and pathological cohorts and fall-short of clinical utilization and application. Specifically, our contributions can be summarized as follows:
\begin{itemize}
\item Our proposed method is unique in the sense that we propose a fully automated system with a geodesic map of bones automatically injected into the deep learning settings unlike the state-of-the-art deep approaches where landmarks are annotated in Euclidean space~\cite{zhang_2017,zhang_october_2017}.
\item While other works learn landmark locations using only spatial location of the landmarks in a digitized image (on the grid) along with context information, we propose to learn the sparsely-spaced landmark relationship on the same bone by utilizing a U-Net based landmark detection algorithm. Then, an LSTM based learning algorithm is developed to identify closely-spaced landmarks. We consider the landmarks as states of the LSTM, and operate completely on the geodesic space. This approach is not only realistic, but it is also computationally more feasible.
\item We present in-depth analysis of architecture design parameters such as the effect of growth rate in segmentation, the use of different pooling functions both in detection and segmentation tasks, and the harmony of dropout regularization with pooling functions. 
\item Our dataset includes highly variable bone deformities along with other challenges of the CBCT scans. For an extremely challenging dataset, the proposed geodesic deep learning algorithm is shown to be robust by successfully segmenting the mandible bones and providing highly accurate anatomical landmarks. 
\end{itemize}

\section{Methods}
The proposed system for segmentation and landmarking comprises three steps (see Figure~\ref{fig:pipeline} for the overview). Step 1 includes a newly proposed segmentation network for mandible based on a unified algorithm combining U-Net and DenseNET with carefully designed network architecture parameters and a large number of layers (called Tiramisu). In Step 2, we propose a U-Net based geodesic learning architecture to learn true and more accurate spatial relationships of anatomical landmarks on the segmented mandible. Finally, in Step 3, we identify closely-spaced landmark locations by a classification framework where we utilize an LSTM network. 

\subsection{Step 1: Segmentation Network}
Recently, CNN based approaches such as U-Net~\cite{unet2015}, fully convolutional network (FCN)~\cite{fcn}, and encoder-decoder CNNs~\cite{cardiacnet} have achieved increasing success in image segmentation. 
These methods share the same spirit of obtaining images at different resolutions by consecutive downsampling and upsampling to make pixel level predictions. Despite the significant progress made by such standard approaches towards segmentation, they often fail to converge in training when faced with objects with high variations in shape and/or texture, and complexities in the structure. 
Another challenge is the optimization of massive amount of hyper-parameters in deep nets. Inspired by the recently introduced notion of densely connected networks (DenseNET) for object recognition~\cite{denseNet}, a new network architecture was presented by J{\'{e}}gou et al.~\cite{tiramisu} for  semantic segmentation of natural images, called Fully Convolutional DenseNET (or \textit{Tiramisu} in short). In this study, we adapt this Tiramisu network for medical image segmentation domain through significant modifications: 

(1) We replaced all default pooling functions (often they are defined as max pooling) with average pooling to increase pixel-level predictions. Although pooling functions in the literature have been reported to perform similarly in various tasks, we hypothesize that average pooling is more suitable for pixel level predictions. Because, average pooling identifies the extent of an object while max-pooling only identifies the discriminative part.  

(2) We explored the role of dropout regularization on segmentation performance with respect to the commonly used batch normalization (BN) and pooling functions. Literature provides mixed evidence for the role of these regularizers.

(3) We investigated the effect of growth rate (of dense block(s)) on the segmentation performance. While a relatively small growth rate has been found successful in various computer vision tasks, the growth rate of dense blocks is often fixed and its optimal choice for segmentation task has not yet been explored. 

(4) We examined appropriate regularization as well as network architecture parameters, including number of layers, to avoid the use of post-processing methods such as CRF (conditional random field). It is common in many CNN-based segmentation methods to use such algorithms so that the model predictions are further refined because the segmentation accuracy is below an expected range. 

\begin{figure}[t]
\centering
\begin{subfigure}{0.48\linewidth}
	\includegraphics[width=\textwidth]{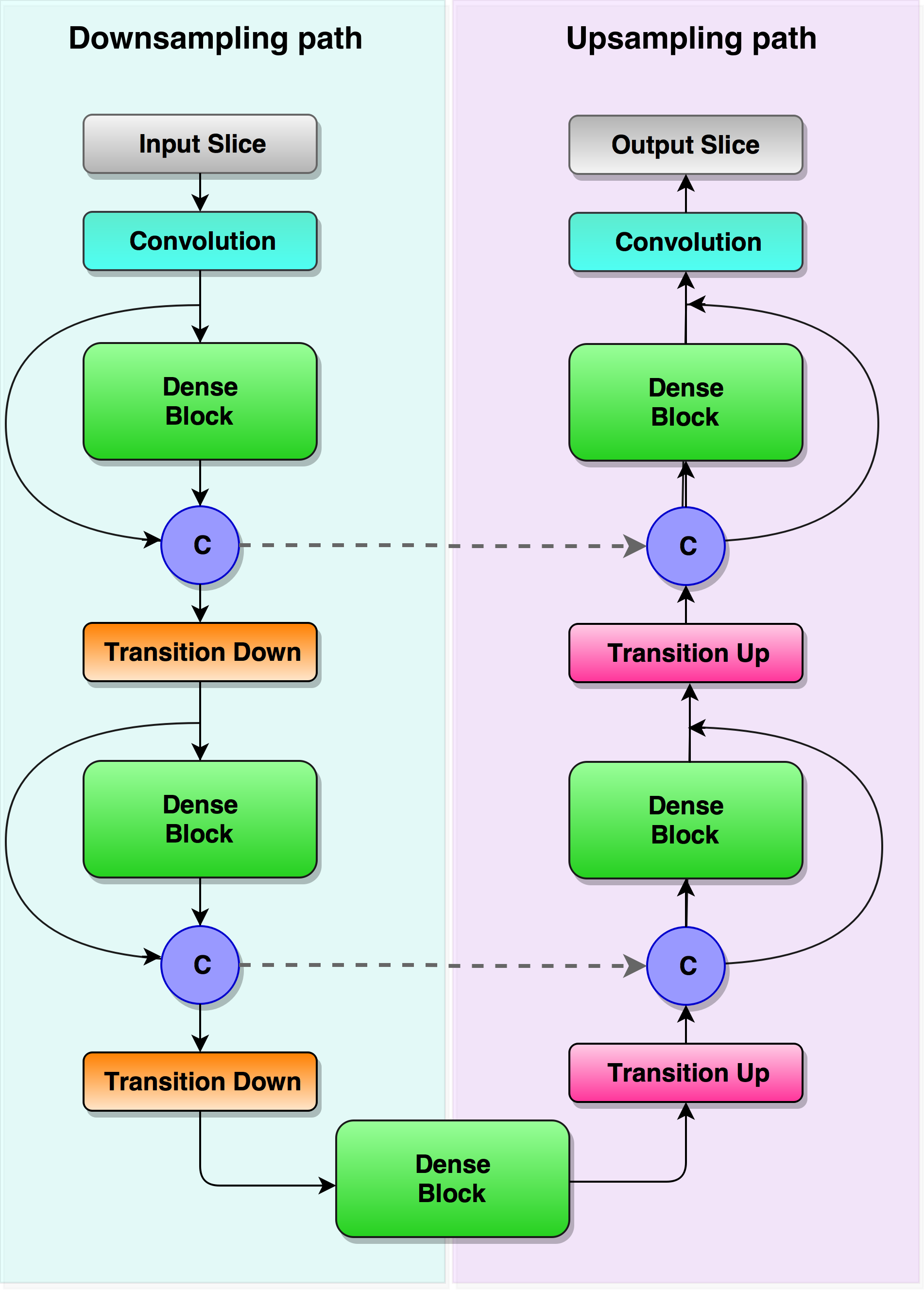}
\caption{Fully Convolutional DenseNet with 103 layers.}
\end{subfigure}
\begin{subfigure}{0.42\linewidth}
	\raisebox{2mm}{\includegraphics[width=\textwidth]{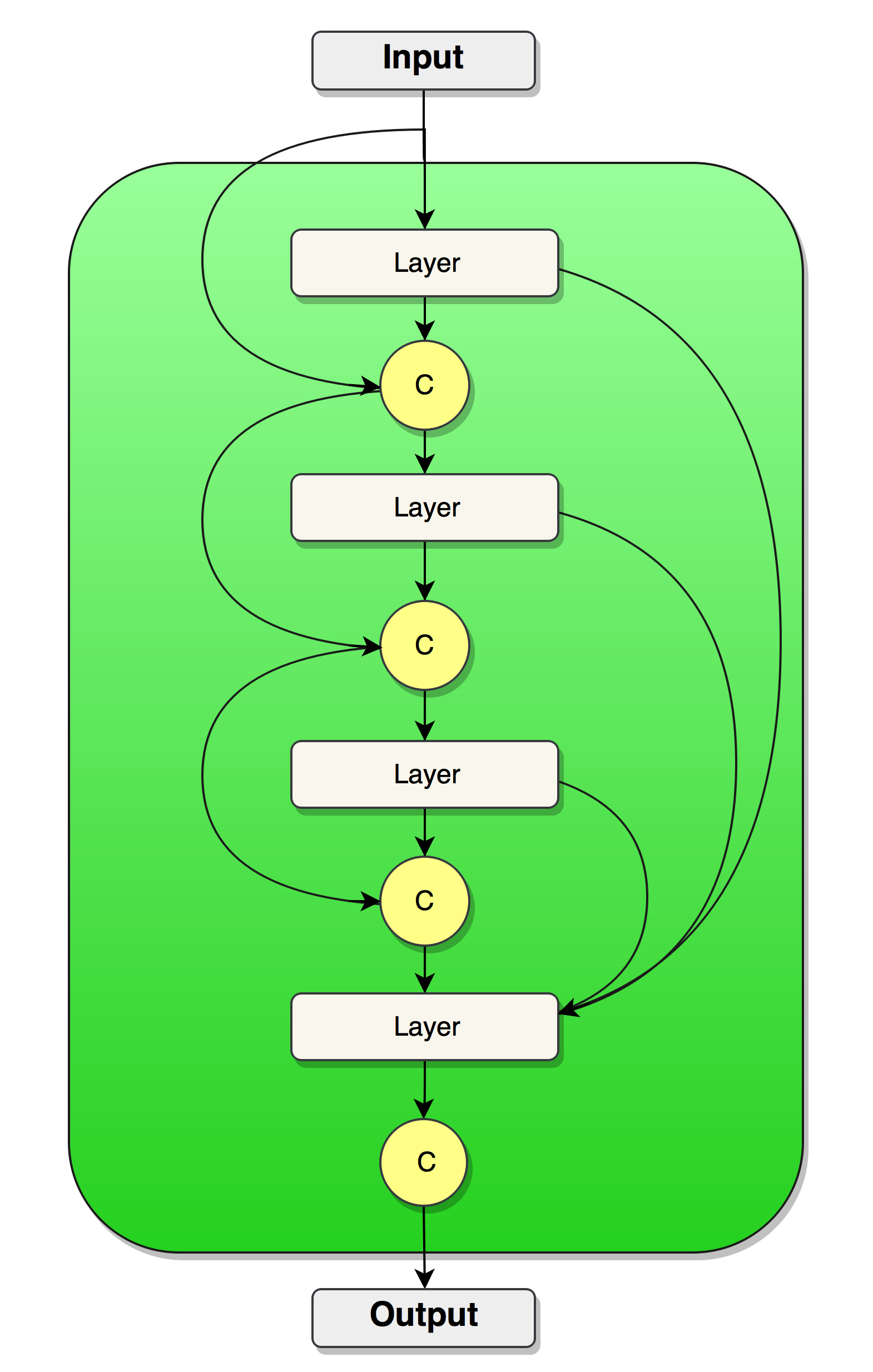}}
	\caption{ Content of a dense block.}
\end{subfigure}
\caption{(a) General architecture of the Tiramisu~\cite{tiramisu} is illustrated. The architecture is composed of downsampling and upsampling paths including Convolution,  Dense Block, Concatenation (C), Skip Connection (dashed lines), Transition Down, and Transition Up layers. Concatenation layer appends the input of the dense block layer to the output of it. Skip connection copies the concatenated feature maps to the upsampling path. (b) A sample dense block with 4 layers is shown to its connections. With a growth rate of $k$, each layer in dense block appends \textit{k} feature maps to the input. Hence, the output contains $4 \times k$ features maps.\label{fig:denseNetArchitecture}}
\end{figure}


Figure 5 illustrates the Tiramisu network architecture (a) and the content of a dense block (b), respectively. Tiramisu network is extremely deep, including 103 layers~\cite{tiramisu} as compared to the U-Net which has only 19 layers in our implementation. The input of the Tiramisu network was the 2D sagittal slices of the CBCT scan of patients with CMF deformities, and the output was the binary 2D sagittal slices with the mandible segmented (see Figure 4 for an example workflow of a 2D slice). The architecture consisted of 11 dense blocks with 103 convolutional layers. Each dense block contained a variable length of layers and the growth rate was set specifically for each dense block based on extensive experimental results and comparison. The network was composed of approximately $9M$ trainable parameters. We trained the revised Tiramisu from scratch without the need for data augmentation and complex post-processing. Details of the network parameters are given in Tables~\ref{table:DenseNet103} and~\ref{table:tiramisuParams}.



\begin {table}[h]
\centering
\caption{The network architecture of the Tiramisu segmentation engine.}
\label{table:DenseNet103}
\begin{tabular}{|l|c|}
\hline
 \textbf{Layers applied} &
\textbf{\# of feature maps}\\ \hline
Input & 1 \\ \hline
$3 \times 3$ Convolution & 48 \\ \hline
Dense Block (4 layers) + Transition Down & 112 \\ \hline
Dense Block (5 layers) + Transition Down & 192 \\ \hline
Dense Block (7 layers) + Transition Down & 304 \\ \hline
Dense Block (10 layers) + Transition Down & 464 \\ \hline
Dense Block (12 layers) + Transition Down & 656 \\ \hline
Dense Block (15 layers) & 896 \\ \hline 
Transition Up + Dense Block (12 layers)& 1088 \\ \hline 
Transition Up + Dense Block (10 layers)& 816 \\ \hline 
Transition Up + Dense Block (7 layers)& 578 \\ \hline 
Transition Up + Dense Block (5 layers)& 384 \\ \hline 
Transition Up + Dense Block (4 layers)& 256 \\ \hline 
$1 \times 1$ Convolution & 2 \\ \hline 
Softmax & 2 \\
\hline
\end{tabular}
\end{table}

\begin {table}[h]
\captionsetup{labelfont={color=black},font={color=black}}
\centering
\caption{The network architecture parameters of the Tiramisu segmentation engine}
\label{table:tiramisuParams}
\begin{tabularx}{0.8\columnwidth}{@{\extracolsep\fill}lcr}
\toprule
\textbf{Hyper-Parameters} & \textbf{Value} \\
\midrule
Learning-Rate & 0.00005 \\
Drop-out & 0.2 \\
Network Weight Initialization & Xavier Initializer\\
Bias Initializer & Zero Initializer \\
Activation Function & Relu \\
Growth Rate & 24 \\
Normalization & Batch Normalization \\
\toprule
\textbf{Network Parameters} & \textbf{Value} \\
\midrule
Pooling & Average \\
Batch-Size & 3 \\
Optimization & Adam\\
\bottomrule
\end{tabularx}
\end{table}

\subsection{Step 2: Geodesic Learning for Landmarking}
We approach the problem of anatomical landmarking (i.e. landmark detection) as a learning problem. The state-of-the-art method in the literature, proposed by Zhang et al., adopts a U-Net architecture to learn the locations of the anatomical landmarks~\cite{zhang_2017}. For a given 3D CBCT scan $\mathbf{X}$ and a landmark $l$, authors~\cite{zhang_2017} created three displacement maps $\mathbf{D}^{l,x}, \mathbf{D}^{l,y}, \mathbf{D}^{l,z}$  corresponding to $x,y$, and $z$ axes~\cite{zhang_2017}. That is, if there are $N_l$ landmarks, $N_l\times 3$ displacement maps are generated. Displacement maps,  also called \textit{heatmaps}, were created using a  simple Euclidean metric measuring the distance of a landmark to a reference point (i.e., (0,0) index of image). Although the method is simple to implement and efficient within the multi-task learning platform, it does not incorporate information about the object of interest (i.e. mandible) and works on the image space. In addition, on account of generating $3$ channel redundant heatmap for each landmark, the method generates a large number of heatmaps when the number of landmarks is high. Lastly, the method operates directly on the Euclidean space and it does not capture the underlying data distribution, which is non-Euclidean in nature.

To alleviate these problems and to solve the landmarking problem directly on the shape space, we propose to use a Geodesic Distance Transform to learn the relationship of landmarks directly on the shape space (mandible surface). To this end, we first apply linear time distance transform (LTDT)~\cite{ltdt} to the segmented mandible images (i.e., binary) and generate signed distance maps. Assuming $\mathbf{I}$ is a 3D-segmented binary image (mandible) obtained at Step 1 from a given CBCT scan $\mathbf{X}$ in the domain $\Omega=\{1,...,n\} \times \{1,...,m\}$, Mandible \textit{M} is represented by all white voxels (i.e., $I(v)=1$), while Mandible complement (background) $M^\textit{C}$ is represented by all black voxels (i.e., $I(v)=0$)~\cite{ltdt_2}:

\begin{equation}
\begin{aligned}
	\mathbf{M} = \{v \in \Omega | I(v)=1\} \\
    \mathbf{M^\textit{\textbf{C}}} = \{v \in \Omega | I(v)=0\}.
\end{aligned}
\end{equation}

LTDT represents a map such that each voxel $v$ is the smallest Euclidean distance from this voxel to the $\textbf{M}^\textit{\textbf{C}}$:

\begin{equation}
\begin{aligned}
	LTDT(v) = 
    min\{dist(v,q) | q \in \textbf{M}^\textit{\textbf{C}}\}.
\end{aligned}
\end{equation}

Then, the signed LTDT (i.e., sLTDT) of $\mathbf{I}$ for a voxel $v$ can be represented as:

\begin{equation}
\begin{aligned}
sLTDT(v)=
\begin{cases}
	LTDT(v) & \text{if }v \in \textbf{M} \\
    -min\{dist(v,q) | q \in \textbf{M}\}  & \text{if }v \in \textbf{M}^\textit{\textbf{C}}.
\end{cases}
\end{aligned}
\end{equation}




\noindent For each landmark $l$, we generate a geodesic distance map $\mathbf{D}^{G}_l$. To do so, we find the shortest distance between landmark $l$ and each voxel $v$ as:

\begin{equation}
\begin{aligned}
   \mathbf{D}_l^G(v) = 
   \begin{cases}
    	\min \pi(l,v) & \text{if } v\in \textbf{M} \\
        \inf & \text{if }v \in \textbf{M}^\textit{\textbf{C}},
    \end{cases}
\end{aligned}
\end{equation}


\noindent where $\pi$ indicates all possible paths from the landmark $l$ to the voxel $v (v \in \textbf{M})$ . Since the shortest distance between two points is found on the surface (i.e., mandible), it is called geodesic distance~\cite{geodesicPaths, Datar2013} as a convention. To find the shortest path $\pi$, we applied Dijkstra's shortest path algorithm. For each landmark $l$, we generated one geodesic map as $\mathbf{D}_l^G$. For multiple landmarks, as is the case in our problem, we simply combine the geodesic maps to generate one final heat map, which includes location information for all landmarks. Final geodesic map for all landmarks is obtained through hard minimum function $\mathbf{D}_{\mathbf{I}}^G=\min_{} (\mathbf{D}_{l_{1}}^G \circ \mathbf{D}_{l_{2}}^G  \circ ... \circ \mathbf{D}_{l_n}^G )$, where $\circ$ indicates pixel-wise comparison of all maps. In other words, the final geodesic map $\mathbf{D}_{\mathbf{I}}^G$ includes $n$ extrema (i.e., minimum) identifying the locations of the $n$ landmarks.  

\color{black}

To learn the relationship of $n$ landmark points on the mandible, we design a landmark localization network, based on the Zhang's U-Net architecture~\cite{zhang_2017}. Tiramisu network could perhaps be used for the same purpose. However, the data was simplified in landmark localization due to geodesic distance mapping, and Zhang's U-Net uses only 10\% of the overall parameter space for landmark localization. The improved Zhang's U-Net accepts $2D$ slices of the signed distance transform of the segmented mandible ($\mathbf{I}$) as the input, and produces the $2D$ geodesic map ($\mathbf{D}_{\mathbf{I}}^G$) revealing the location of $N_l$ landmarks as the output. The details of the landmark localization architecture (improved version of the Zhang's U-Net) with 19 layers and parameters are given in Tables~\ref{table:unet} and~\ref{table:unetParams} respectively. Briefly, the encoder path of the U-Net was composed of 3 levels. Each level consisted of (multiple) application(s) of convolutional nodes: $5 \times 5$ convolutions, batch normalization (BN), rectified linear unit (ReLU), and dropout. Between each level max pooling, downsampling with a stride of 2, was performed. The number of features obtained at the end of each level in the encoder path were $32$, $64$, and $128$ respectively. Similar to the encoder path, the decoder path was also composed of $3$ levels. In contrast to encoder path, dropout was not applied in the decoder path. Between the levels in the decoder path, upsampling operation was applied. To emphasize the high-resolution features that may be lost in the encoder path, copy operation was used in the decoder path. Copy operation, as the name implies, concatenated the features at the same $2D$ resolution levels from the encoder path to the decoder path. The number of features obtained at the end level in the decoder path was $64$, $32$, and $2$ respectively. We have chosen the optimization algorithm as RMSProp~\cite{rmsprop} due to its fast convergence and adaptive nature. The initial learning rate was set to 1e-3 with an exponential decay of 0.995 after each epoch (Table~\ref{table:unetParams}). At the end of the decoder path, softmax cross entropy was applied. We preferred using softmax cross entropy as the loss function rather than the mean squared error (MSE) due to serious convergence issues in the MSE loss. Hence, we quantized the geodesic map in the range $[0-20]$, where the limit $20$ was set empirically. The network was composed of $\approx 1M$ trainable parameters. Compared to the Zhang's U-Net~\cite{zhang_2017}, in our improved implementation, in addition to the $5 \times 5$ convolutions, on the expanding path at level 2, we kept the symmetry in the number of features obtained as in the contracting path. These alterations made sure Zhang's U-Net to work without failures. 

\begin {table}[h]
\centering
\caption{The network architecture of the \underline{improved} Zhang's U-Net for sparsely-spaced landmarks}
\label{table:unet}
\begin{tabular}{|l|c|c|}
\hline
\parbox[c]{3cm}{\centering \textbf{Layers applied}} & \parbox[c]{2cm}{\centering \textbf{Slice Size}} &
\parbox[c]{2.5cm}{\centering \textbf{Number of feature}\\ \textbf{maps}}\\
\hline
\hline
Input & $256 \times 256$ & 1 \\ \hline
$5 \times 5$ Convolution & $256 \times 256$ & 32 \\ \hline
$5 \times 5$ Convolution & $256 \times 256$ & 32 \\ \hline
\hline
Max-pooling & $128 \times 128$ & 32 \\ \hline
\hline
$5 \times 5$ Convolution & $128 \times 128$ & 64 \\ \hline
$5 \times 5$ Convolution & $128 \times 128$ & 64 \\ \hline
\hline
Max-pooling & $64 \times 64$ & 64 \\ \hline
\hline
$5 \times 5$ Convolution & $64 \times 64$ & 128 \\ \hline
$5 \times 5$ Deconvolution & $64 \times 64$ & 64 \\ \hline
\hline
Upsampling + Copy & $128 \times 128$ & 128 \\ \hline
\hline
$5 \times 5$ Deconvolution & $128 \times 128$ & 64 \\ \hline
$5 \times 5$ Deconvolution & $128 \times 128$ & 32 \\ \hline
\hline
Upsampling + Copy & $256 \times 256$ & 64 \\ \hline
\hline
$5 \times 5$ Deconvolution & $256 \times 256$ & 32 \\ \hline
$5 \times 5$ Deconvolution & $256 \times 256$ & 32 \\ \hline
$5 \times 5$ Deconvolution & $256 \times 256$ & 21 \\ \hline
\hline
Softmax & $256 \times 256$ & 21 \\

\hline
\hline
\end{tabular}
\end{table}

\begin {table}[h]
\captionsetup{labelfont={color=black},font={color=black}}
\centering
\caption{The network architecture parameters of the \underline{improved} Zhang's U-Net for sparsely-spaced landmarks}
\label{table:unetParams}
\begin{tabularx}{0.8\columnwidth}{@{\extracolsep\fill}lcr}
\toprule
\textbf{Hyper-Parameters} & \textbf{Value} \\
\midrule
Learning-Rate & 1e-3 \\
Decay-Rate & 0.995 \\
Drop-out & 0.2 \\
Network Weight Initialization & Xavier Initializer \\
Bias Initializer & Zero Initializer \\
Normalization & Batch Normalization \\
Pooling & Maxpool \\
Batch-Size & 3 \\
Optimization & RMSProp\\
\bottomrule
\end{tabularx}
\end{table}


\subsection{Step 3: Localization of Closely-Spaced Landmarks}
Fusion of geodesic maps through pixel-wise hard-coded minimum function is reliable when landmarks are sufficiently distant from each other. In other words, if landmarks are very close to each other, then the combined geodesic map $\mathbf{D}_{\mathbf{I}}^G$ may have instabilities in locating its extrema points. In particular for our case, it was not possible to localize specifically ``Menton'' and other mid-sagittal closely-spaced landmarks in a clinically acceptable error range (i.e., $\leqslant 3mm$). In order to avoid such scenarios, we propose to divide the landmarking process into two distinct cases: learning closely-spaced and sparsely-spaced landmarks separately. First, we divide the mandible landmarks into sparsely and closely-spaced sets. Sparsely-spaced landmarks (N=5) were defined in the inferior, superior-posterior-left, superior-posterior-right, superior-anterior-left, and superior-anterior-right regions. Closely-spaced landmarks (N=4) were defined as the ones that are closely tied together (Infradentale $(Id)$, B point $(B)$, Pogonion $(Pg)$, and Gnathion $(Gn)$). 

\begin{figure}[t]
\centering
\includegraphics[width=\linewidth,valign=t]{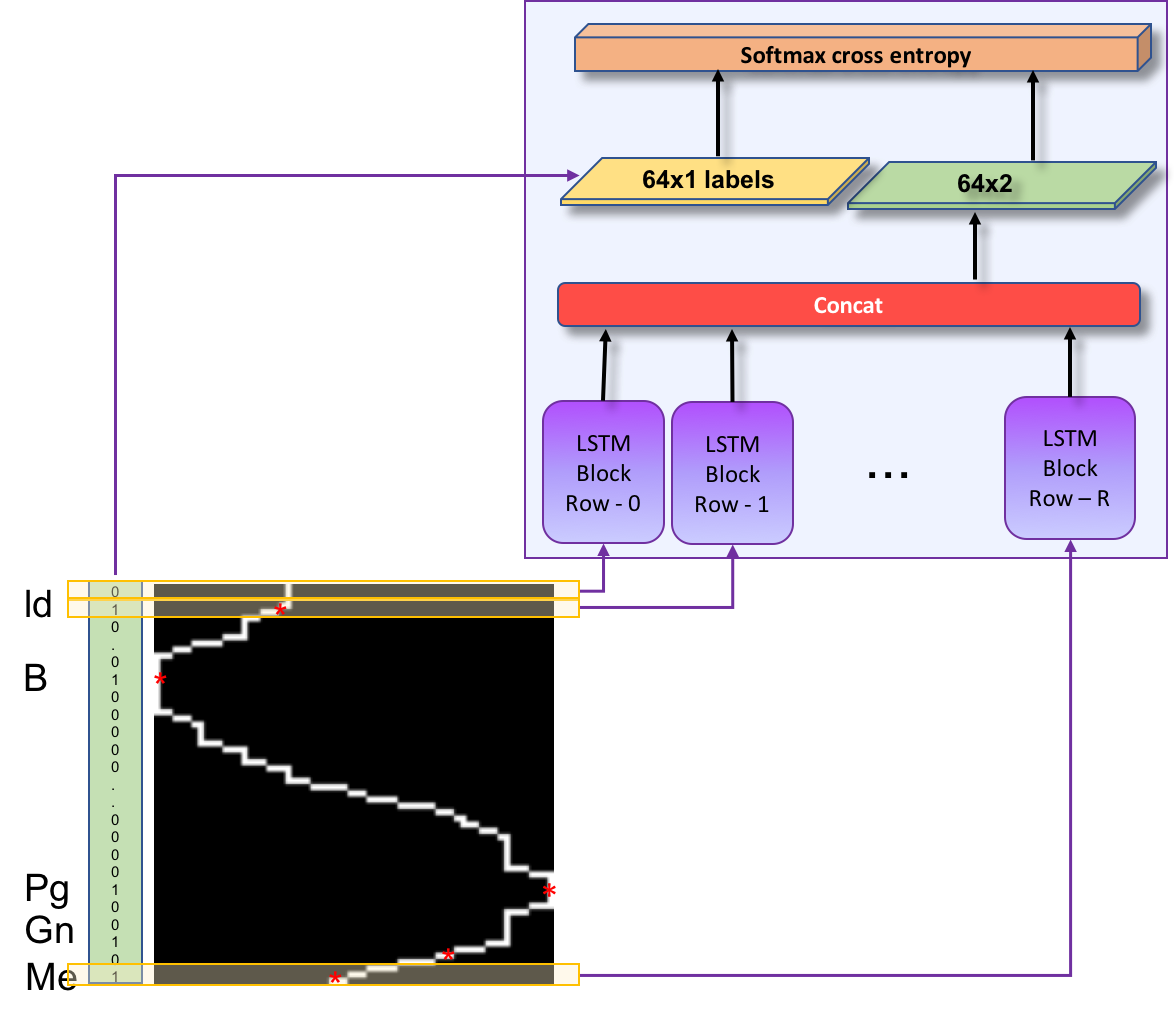}
\caption{LSTM network input-outputs. Each row of the scaled sagittal boundary image is input to the corresponding LSTM block, and binary $1D$ vector of locations annotated as landmark ($1$), or no-landmark ($0$) is output.}
\label{fig:lstm_input}
\end{figure}
\color{black}

\begin{figure}[h]
\centering
\begin{subfigure}{0.60\linewidth}
	\includegraphics[height=5cm]{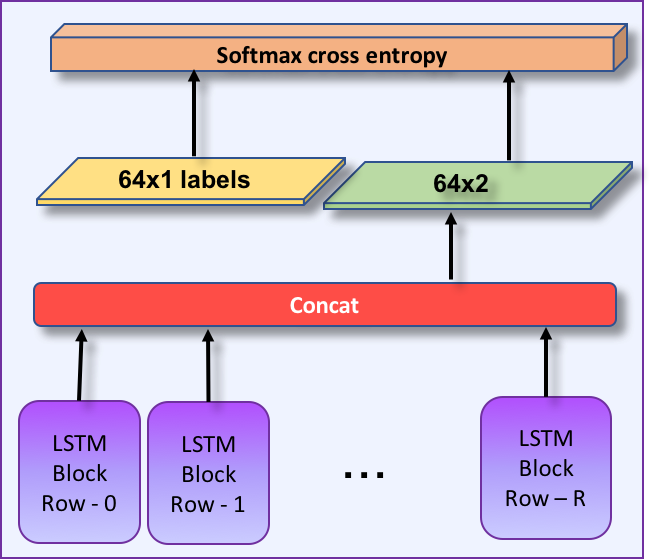}
	\caption{LSTM Network}
    \label{fig:denseBlock}
\end{subfigure}
\qquad
\begin{subfigure}{0.30\linewidth}
	\includegraphics[height=5cm]{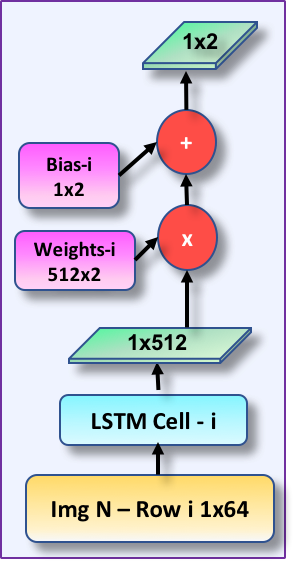}
	\caption{LSTM Block}
    \label{fig:denseNetArchitecture}
\end{subfigure}
\caption{Details of the network architecture (LSTM) for identifying closely-spaced landmarks. Gnathion $(Gn)$, Pogonion $(Pg)$, B Point $(B)$, and Infradentale $(Id)$ are determined once the Menton $(Me)$ is detected through U-Net architecture as shown in Step 3 of the Figure~\ref{fig:pipeline}. Input image resolution is RxK, and the LSTM cell is composed of 512 hidden units.}
\label{fig:lstmNetwork2}
\end{figure}

Note that these anatomical landmarks often reside on the same sagittal plane in the same order according to the mid-point of the lower-jaw incisors. We propose to capture this order dependence by using an LSTM architecture in the sagittal axis of the images containing the landmark ``Menton'' (Figure~\ref{fig:lstm_input}). The rationale behind this choice is that LSTM network is a type of RNN introduced by Hochreiter et al.~\cite{Hochreiter:1997:LSM:1246443.1246450} in 1997, modeling the temporal information of the data effectively. Although the imaging data that we used for landmark localization does not include temporal information in the standard sense, we modeled the landmark relationship as a temporal information due to their close positioning in the same plane. This phenomenon is illustrated in Figure~\ref{fig:lstm_input}. The input data to the LSTM network was a $64 \times 64$ mandible binary boundary image of the sagittal plane of the landmark $Me$, and the output is a vector of 0's and 1's: while 0 refers to non-landmark location, 1 refers to a landmark location in the sagittal axis. Figure~\ref{fig:lstmNetwork2} shows further details of the LSTM network and content of a sample LSTM block that we used for effective learning of closely-spaced landmarks.

To generate the training data, the sagittal slice region containing the closely-spaced landmarks ``Menton'', ``Gnathion'', ``Pogonion'', ``B-point'' and ``Infradentale'' is scaled into a binary boundary image of size $64 \times 64$. The $5$ landmark locations (marked by red circles in Figure~\ref{fig:lstm_input}) on this boundary image are parametrized as (\textbf{\textit{x}},\textbf{\textit{y}}), where \textbf{\textit{y}} is the row number in the range $0$ to $64$, and \textbf{\textit{x}} is the white boundary column number of the corresponding row \textbf{\textit{y}}.  Due to having $40$ training patients' scans, there are only $40$ sagittal slices containing the closely-spaced landmarks, which is insufficient for the LSTM training. Hence, in order to increase the number training images, we performed Principal Component Analysis (PCA).

LSTM network is composed of 64 cells (as demonstrated in Figure~\ref{fig:lstmNetwork2}), and each cell in the LSTM network consisted of $512$ units. The training images were row-wise input to the LSTM network such that $n^{th}$ row was input to the corresponding $n^{th}$ cell of the network. The output of each cell was multiplied by $512 \times 2$ weight and $1 \times 2$ bias was added. The resultant $1 \times 2$ tensors at each cell were concatenated and softmax cross entropy was applied as a loss function. 


\begin{figure*}[ht]
\begin{subfigure}{0.33\linewidth}
    \includegraphics[width=\textwidth]{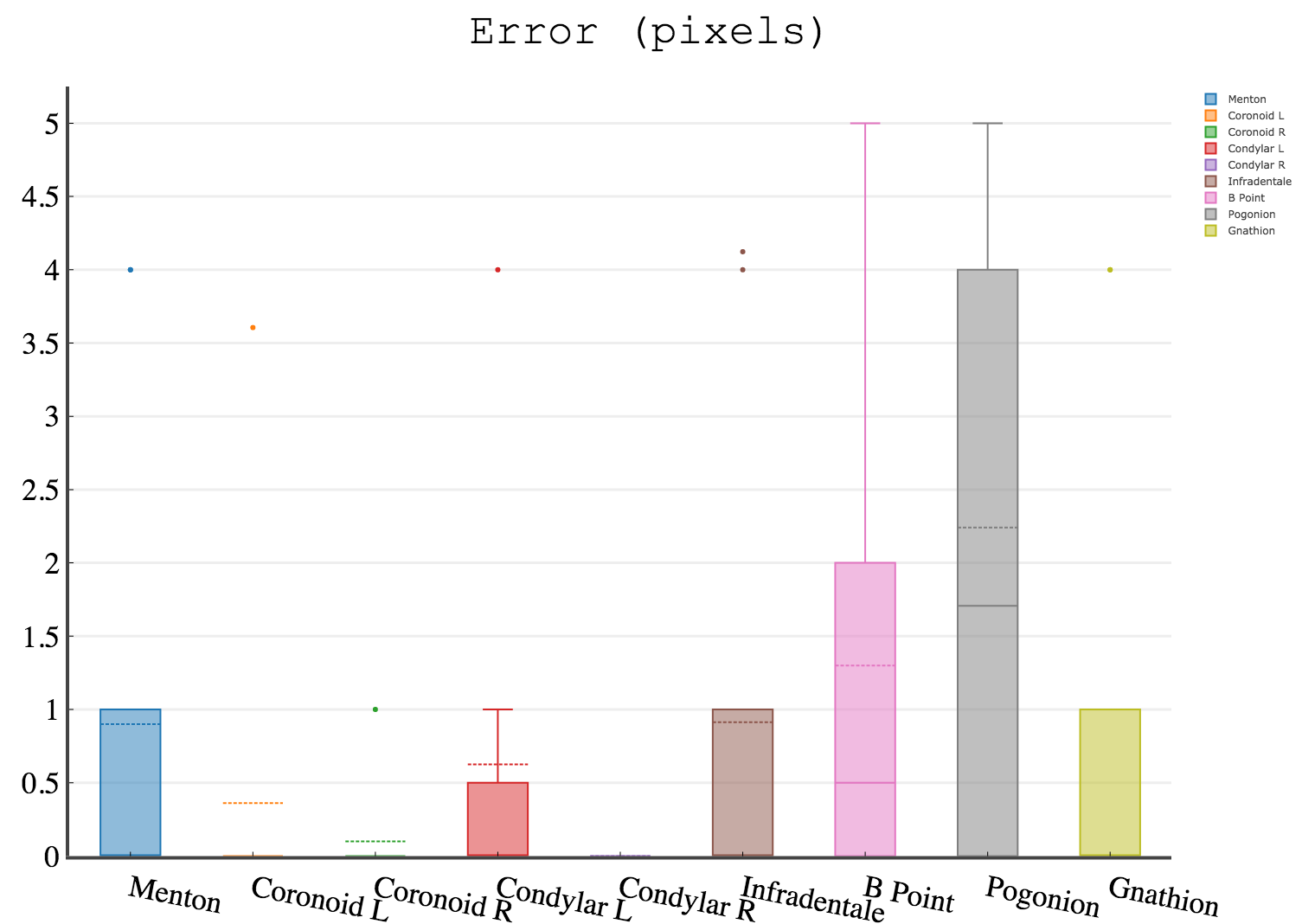}
	\caption{Pixel Space Errors}
    \label{fig:errPixelSpace}
    \vspace*{2mm}
\end{subfigure}
\begin{subfigure}{0.32\linewidth}
    \includegraphics[width=\textwidth]{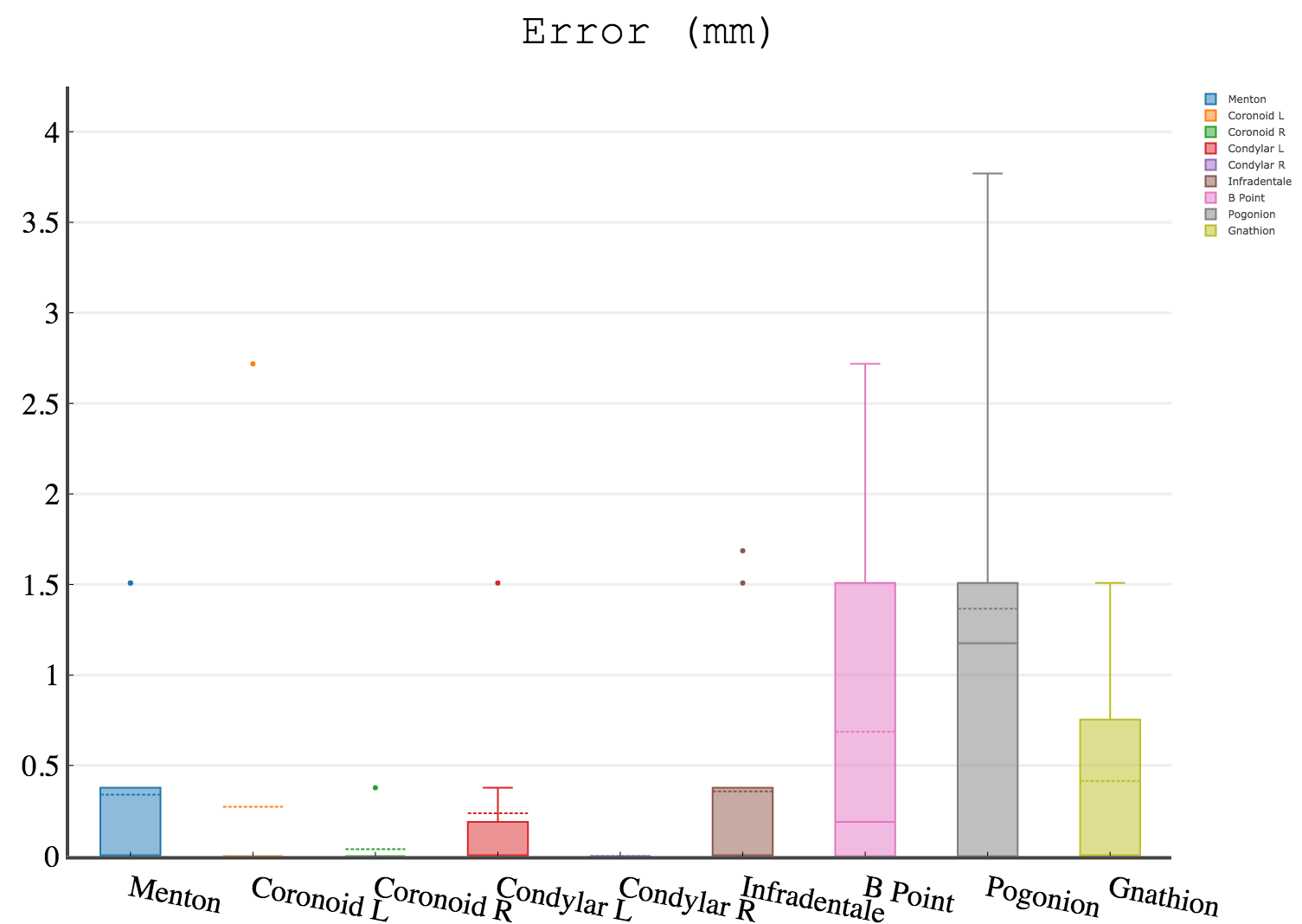}
	\caption{Volume Space Errors}
    \label{fig:errRasSpace}
\end{subfigure}\qquad
\begin{subfigure}{0.32\linewidth}
\includegraphics[width=\textwidth]{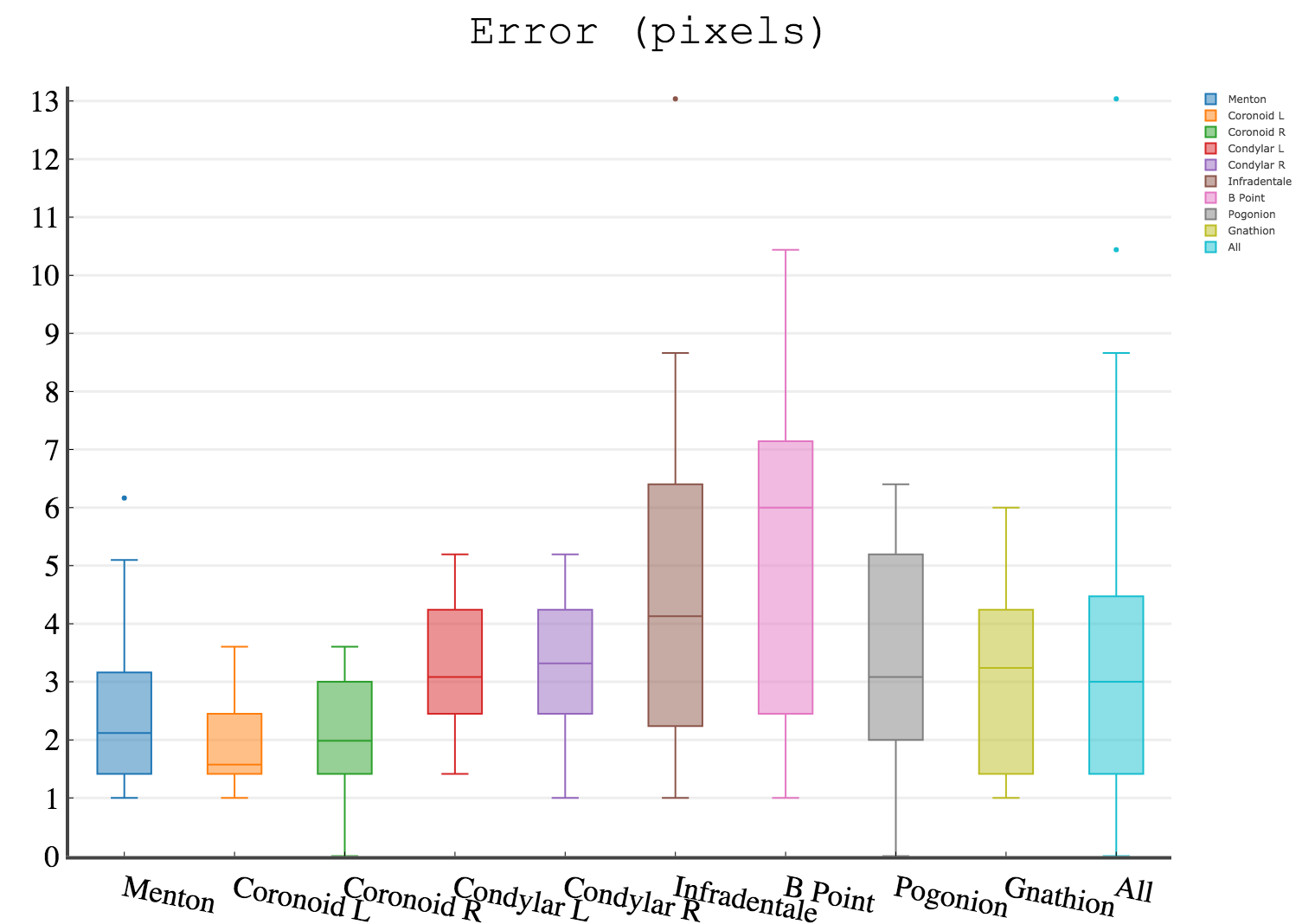}
\caption{Expert Reading Variations}
\label{fig:expertVariation}
\end{subfigure}
\label{fig:errorPlots}
\caption{(a) Errors in pixel space, (b) errors in the volume space, (c) inter-observer reading variations in pixel space. \label{fig:lmerrors}}
\end{figure*}

\subsection{Training Framework: End-to-end vs. Sequential vs Mixed}
Since the proposed learning system is complex, it is worth to explore whether gradient-descent learning system can be applied to the system as a whole (called \textit{end-to-end}). For this purpose, first, we evaluated the performances of each network individually, so named \textit{sequential training} followed by an engineering approach for concatenation of the three networks.  Since end-to-end learning systems require all modules of the complex system to be differentiable, our proposed system was not fully eligible for this learning type. It is because the $3^{rd}$ module (LSTM network for closely-spaced landmark localization) in our proposed system has differentiability issues for the loss function. Therefore, we trained the first and second modules in an \textit{end-to-end} manner while integrating the third network module into this system sequentially. In summary, we devised two alternative methods to solve our overall goal: in the first solution, the overall system was considered as a sequential system. In the second solution, the first two modules of the system were trained in an end-to-end manner with the inclusion of the third module as a sequential block. Owing to the usage of sequential and end-to-end frameworks together, we named the second solution as ``mixed''.

Although end-to-end networks are conceptually and mathematically beautiful, it has a strict condition that each module should be differentiable with respect to the loss function so that a positive impact can be obtained on the final objective. However, as stated in~\cite{end_2_end} and~\cite{limits_end_2_end}, when some modules are not differentiable (as the third module of our proposed method), or when the system is too complex with sparse modules, the overall results may be inferior compared to the sequential method. Due to the differentiability issue in the third module, our system falls into this category. That is, the input to the $3^{rd}$ module is the $2D$ sagittal slice containing the anatomical landmark ``Menton''. Since not every output slice in the training can be used for the $3^{rd}$ module, differentiability is lost. In addition, we observed that unless the first two modules are close to the converging state, it is not possible to localize ``Menton'' more precisely than a random guess. Due to the requirement of convergence within this module,  
eventually, it was not possible to apply LSTM training in a truly end-to-end manner.




\section{Experiments and Results}
\subsection{Data description:} Anonymized CBCT scans of 50 patients (30 female and 20 male, mean age = 22.4 years, standard deviation = 9.6 years) were included in our analysis through an IRB-approved protocol and data sharing agreement. These patients had craniofacial congenital birth defects, developmental growth anomalies, trauma to the CMF, surgical intervention, and included pediatric and adult patients. All images were obtained on a CB MercuRay CBCT system (Hitachi Medical Corporation, Tokyo, Japan). The 12-inch field of view was required for this study to capture the entire length of the airway and was scanned at 10 mA and 100 Kvp. The equivalent radiation dosage for each scan was approximately 300 mSv. After the study had begun, the machine was modified to accommodate 2 mA for the same 12-inch field of view, thus lowering the equivalent radiation dosage for each scan to approximately 132.3 mSv. Each patient's scan was re-sampled from $512 \times 512 \times 512$ to $256 \times 256 \times 512$ to reduce computational cost. In-plane resolution of the scans was noted either as $0.754 mm \times 0.754 mm \times 0.377 mm$ or $0.584 mm \times 0.584 mm \times 0.292 mm$.

Additionally, we tested and evaluated our algorithm(s) using the MICCAI Head-Neck Challenge 2015 dataset~\cite{head_neck_challange_dataset}. MICCAI Head-Neck Challenge 2015 dataset is composed of manually annotated CT scans of $48$ patients from the Radiation Therapy Oncology Group (RTOG) $0522$ study (a multi-institutional clinical trial led by Dr Kian Ang~\cite{ang}). 
For all data, the reconstruction matrix was $512 \times 512$ pixels. The in-plane pixel spacing was isotropic, and varied between $0.76mm \times 0.76mm$ and $1.27mm \times 1.27mm$. The range of the number of slices of the scans were $110$-$190$. The spacing in the z-direction was between $1.25mm$ and $3mm$~\cite{head_neck_challange_dataset}. In the challenge, there were three test results provided, where test data part $1$ (off-site data) and part $2$ (on-site data) did not have publicly available manual annotations to compare to our performances. Hence, we compared our test results to the the cross-validation results as provided in~\cite{miccai_2015_challange_winner}.

\underline{Training deep networks:} We have trained our deep networks with $50$ patients' volumetric CBCT scans in a $5$-fold cross validation experimental design. Since each patient's scan includes $512$ slices (i.e., $2D$ images with $256 \times 256$ pixels in-plane), we had a total of $25,600$ images to train and test the proposed system. In each training experiment, we have used $20,480$ $2D$ images to train the network while the remaining slices ($5,120$) were used for testing. This procedure was repeated for each fold of the data, and average of the overall scores were presented in the following subsections.

\subsection{Evaluation metrics and annotations:}
Three expert interpreters who were blinded annotated the data (one from the NIH team, two from the UCF team). Inter-observer agreement values were computed based on these three annotations. Later, second and third experts (from the UCF team) repeated their manual landmarking processes (after one month period of their initial landmarking) for intra-observer evaluations. Experts used freely available 3D Slicer software for the annotations. Annotated landmarks were saved in the same format of the original images, where landmark positions in a neighborhood of $3 \times 3 \times 3$ were marked according to the landmark ID while the background pixels were marked as $0$.

A simple median filtering was used to minimize noise in the scans. No other particular preprocessing algorithm was used. Experiments were performed through a 5-fold cross-validation method. Intersection of Union (IoU) metric was used to evaluate object detection performance. For evaluating segmentation, we used the standard DSC (dice similarity coefficient), Sensitivity, Specificity, and HD (Hausdorff Distance) ($100\%$ percentile). As a convention, high DSC, sensitivity, specificity and low HD indicate a good performance. The accuracy of the landmark localization was evaluated using the detection error in pixel space within a $3 \times 3 \times 3$ bounding box.  Inter-observer agreement rate was found to be 91.69\% for segmentation (via DSC).   


\subsection{Evaluation of Segmentation}
The proposed segmentation framework achieved highly accurate segmentation results despite the large variations in the imaging data due to severe CMF deformities. Table~\ref{table:segAccuracies} summarizes the segmentation evaluation metrics and number of parameters used for the proposed and the compared networks. The proposed segmentation network outperformed  the state-of-the-art U-Net~\cite{unet2015}. Specifically, we have improved the success of the baseline U-Net framework by increasing the number of layers into 19. In terms of the dice similarity metric, both improved Zhang's U-Net and the proposed segmentation network were statistically significantly better than the baseline U-Net ($P=0.02$ st-test).
In summary, (i) there is no statistically significant difference noted between our proposed method and the manual segmentation method ($P=0.77$); (ii) there is a statistically significant difference between our proposed method and the baseline U-Net ($p=0.02<0.05$); (iii) there is no statistically significant difference noted between the proposed method and our improvement over the Zhang’s U-Net ($P=0.28$). It is also worth to note that the proposed Tiramisu network performed more robustly in training, converging faster than the improved Zhang's U-Net despite the larger number of parameters in the Tiramisu.




\begin {table}[h]
\centering
\caption{Evaluation of the segmentation algorithms. Higher IoU(\%) and DSC (\%), and lower HD (mm) indicate better segmentation performance. Improved Zhang's U-Net is built on top of Zhang's U-Net implementation~\cite{zhang_2017}.}
\label{table:segAccuracies}
\resizebox{\columnwidth}{!}{
\begin{tabular}{lcccccc}
\hline
Method & IoU   & DSC & HD & Layers & \# of params.\\
\hline
Baseline U-Net~\cite{unet2015}  &   100     &  91.93   & 5.27  & 31 & $\approx$ 50M\\ 
\textbf{Improved Zhang's U-Net}  & 100      & 93.07    & 5.87 & 19 &$\approx$ 1M\\      
\textbf{Proposed (Tiramisu) } & 100      &  \textbf{93.82}    & \textbf{5.47} &103 &  $\approx$ 9M\\ 
\hline
\end{tabular}
}
\end{table}

We evaluated the segmentation performances on different datasets and training styles (sequential vs. mixed learning) and summarized the results in  Table~\ref{table:seg_perf}. 
With the MICCAI Head-Neck Challenge 2015 dataset, we obtained a dice accuracy of $93.86$\% compared to $90$\%~\cite{miccai_2015_challange_winner}. High accuracies of the MICCAI Head-Neck Challenge 2015 and the NIH datasets imply the robustness of the Tiramisu segmentation network. It should be noted that  MICCAI Head-Neck Challenge 2015 dataset contains mainly scans with imaging artifacts as well as different diseases. Closer inspection of Table~\ref{table:seg_perf} also shows that a simple post-processing step such as ``Connected Component Analysis'' and ``3D fill'' were important to decrease the number of the false positives and false negatives in the challenge dataset. The slightly lower performances of mixed training with Tiramisu network for both segmentation and landmark localization can be explained by the increased number of parameters but insufficient dataset size to derive learning procedure as a whole. Sequential learning was sufficient to obtain good results in segmentation, though.



\begin{table}[ht]
\captionsetup{labelfont={color=black},font={color=black}}
\centering
\caption{Segmentation performances in different datasets, training paradigms (mixed vs. sequential), and post-processing algorithms.}
\label{table:seg_perf}
\resizebox{\columnwidth}{!}{%
\begin{tabular}{c|c|c|c|c|c}
\cline{1-6}
  & Post-processing & DSC($\%$) & Sensitivity($\%$) & Specificity($\%$) & HD(mm) \TBstrut\\
\cline{1-6}
\multicolumn{1}{ c| }{\multirowcell{3}{\textbf{Sequential} \\ Tiramisu Segmentation \\ MICCAI $2015$}}
& \multicolumn{1}{ c| }{--} & 92.30 & 86.43 & 99.96 & 5.09 \TBstrut\\
\cline{2-6}
& \multicolumn{1}{ c| }{} & & & & \\[0ex]
& \multicolumn{1}{ c| }{\thead{connected component \\ analysis, $3D$ fill}} & 93.86 & 95.23 & 99.99 & 4.58 \Bstrut\\
\cline{1-6}


\multicolumn{1}{ c| }{\multirowcell{3}{\textbf{Sequential} \\ Tiramisu Segmentation \\ NIH Dataset}}
& \multicolumn{1}{ c| }{--} & 92.61 & 93.42 & 99.97 & 8.80 \TBstrut\\
\cline{2-6}
& \multicolumn{1}{ c| }{} & & & & \\[0ex]
& \multicolumn{1}{ c| }{\thead{connected component \\ analysis, $3D$ fill}} & 93.82 & 93.42 & 99.97 & 6.36 \Bstrut\\
\cline{1-6}


\multicolumn{1}{ c| }{\multirowcell{4}{\textbf{Mixed} \\ Tiramisu Segmentation \\ $\rightarrow$ U-Net Landmark Localization\\ NIH Dataset}} 
& \multicolumn{1}{ c| }{--} & 92.09 & 92.10 & 99.96 & 8.30 \TBstrut\\
\cline{2-6}
& \multicolumn{1}{ c| }{} & & & & \\[0ex]
& \multicolumn{1}{ c| }{\thead{connected component \\ analysis, $3D$ fill}} & 92.28 & 92.10 & 99.96 & 7.11 \Bstrut\\
\cline{1-6}


\multicolumn{1}{ c| }{\multirowcell{4}{\textbf{Mixed} \\ Tiramisu Segmentation \\ $\rightarrow$ Tiramisu Landmark Localization\\ NIH Dataset}} 
& \multicolumn{1}{ c| }{--} & 90.10 & 90.53 & 99.97 & 8.80 \TBstrut\\
\cline{2-6}
& \multicolumn{1}{ c| }{} & & & & \\[0ex]
& \multicolumn{1}{ c| }{\thead{Connected component \\ analysis, $3D$ fill}} & 90.10 & 90.52 & 99.97 & 6.36 \Bstrut\\
\cline{1-6}
\end{tabular}
}
\end{table}

\begin{figure*}[t]
\centering
\begin{subfigure}{0.32\linewidth}
	\includegraphics[width=\textwidth]{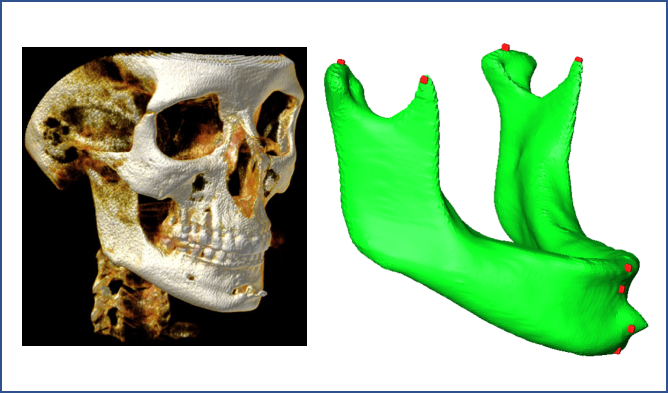}
	\caption{Genioplasty/chin advancement (43 yo)}
    \label{fig:abnormal1}
\end{subfigure}
\begin{subfigure}{0.32\linewidth}
	\includegraphics[width=\textwidth]{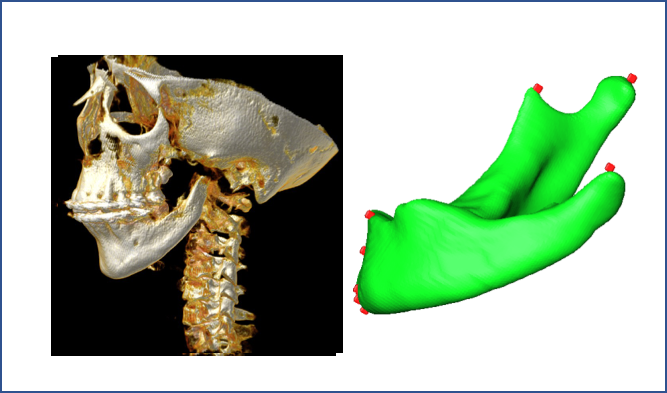}
	\caption{Absent left condyle-ramus unit (15 yo)}
    \label{fig:abnormal2}
\end{subfigure}
\begin{subfigure}{0.32\linewidth}
	\includegraphics[width=\textwidth]{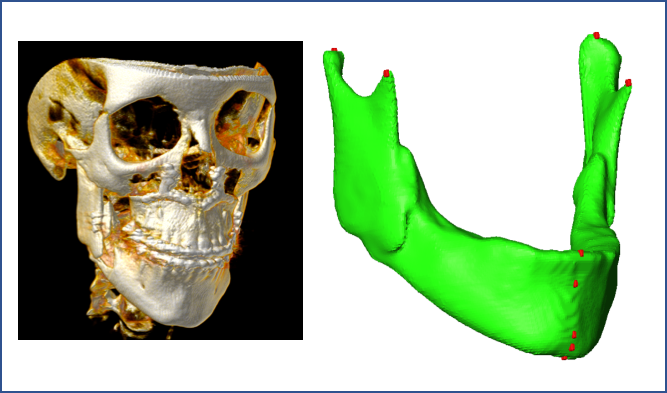}
	\caption{Mandibular implants (17 yo)}
    \label{fig:abnormal3}
\end{subfigure}
 \caption{Experimental renderings demonstrating segmentation and landmark localization results of patients with high anatomical variability due to deformities and surgical intervention}.
 \label{fig:landmarkSkull}
\end{figure*}


\subsection{Evaluation of Landmark Localization}
Ground truth annotations (i.e., manual segmentation and anatomical landmarking) were performed by three experts independently. Inter-observer agreement rate was  $91.69$\%. Figure~\ref{fig:expertVariation} presents per landmark and overall expert reading variations of landmarking in the pixel space. We observed that there was an average $3$ pixel errors among the experts. Hence, any landmarking algorithm leading to error within $3$ pixel range can be considered a clinically acceptable level of success. Figures~\ref{fig:errPixelSpace} and~\ref{fig:errRasSpace} summarize the proposed algorithm's landmark localization errors in the pixel space and the volume space, respectively. 

The mean and median volume space errors for each landmark are presented at Table~\ref{table:regularization}.  For the landmarks Menton, Coronoid Left, Coronoid Right, Condylion Left, Condylion Right, Infradentale, and Gnathion the mean and the median errors in the volume space were less than $1$mm, and for B point and Pogonion the mean and the median errors in the volume space were less than $1.36$ mm. The errors in the pixel space (Figure~\ref{fig:errPixelSpace}) were less than $3$ pixels for all $9$ landmarks, indicating that our method is highly accurate and can be used for clinical applications as it results in less variations than the inter-observer variation rate as explained earlier (Figure~\ref{fig:expertVariation}).


\begin{table}[ht]
\centering
\caption{Landmark localization performances are evaluated for each anatomical landmark on the mandible and with respect to different pooling functions. Errors (in mm) are given both in average (avg) and median (md) values.}
\label{table:regularization}
\resizebox{\columnwidth}{!}{%
\begin{tabular}{cccccc}
\cline{3-6}
 &  & max pool & avg pool & stoc. pool & max pool + \\
 &  &  &  &  & wo drop out \\
\cline{1-6}
\multicolumn{1}{ c| }{\multirow{2}{*}{\textit{$Me$}}} & \multicolumn{1}{ c| }{avg} & 0.33 & 1.35 & 0.37 & 0.03 \\
\multicolumn{1}{ c| }{} & \multicolumn{1}{ c| }{md} & 0 & 0 & 0 & 0 \\
 \hline
 \multicolumn{1}{ c| }{\multirow{2}{*}{\textit{$Cor_L$}}} & \multicolumn{1}{ c| }{avg} & 0.27 & 0.07 & 0 & 0 \\
\multicolumn{1}{ c| }{} & \multicolumn{1}{ c| }{md} & 0 & 0 & 0 & 0 \\
 \hline
 \multicolumn{1}{ c| }{\multirow{2}{*}{\textit{$Cor_R$}}} & \multicolumn{1}{ c| }{avg} & 0.03 & 0.3 & 0.37 & 0.45 \\
\multicolumn{1}{ c| }{} & \multicolumn{1}{ c| }{md} & 0 & 0 & 0 & 0 \\
 \hline
 \multicolumn{1}{ c| }{\multirow{2}{*}{\textit{$Cd_L$}}} & \multicolumn{1}{ c| }{avg} & 1.01 & 0.037 & 0.56 & 0.33 \\
\multicolumn{1}{ c| }{} & \multicolumn{1}{ c| }{md} & 0 & 0 & 0 & 0 \\
 \hline
 \multicolumn{1}{ c| }{\multirow{2}{*}{\textit{$Cd_R$}}} & \multicolumn{1}{ c| }{avg} & 0 & 0.11 & 0.07 & 0.07 \\
\multicolumn{1}{ c| }{} & \multicolumn{1}{ c| }{md} & 0 & 0 & 0 & 0 \\
 \hline
 \multicolumn{1}{ c| }{\multirow{2}{*}{\textit{$Gn$}}} & \multicolumn{1}{ c| }{avg} & 0.41 & 1.64 & 1.35 & 0.49 \\
\multicolumn{1}{ c| }{} & \multicolumn{1}{ c| }{md} & 0 & 0 & 0.18 & 0 \\
 \hline
 \multicolumn{1}{ c| }{\multirow{2}{*}{\textit{$Pg$}}} & \multicolumn{1}{ c| }{avg} & 1.36 & 2.34 & 2.4 & 1.54 \\
\multicolumn{1}{ c| }{} & \multicolumn{1}{ c| }{md} & 1.17 & 0.75 & 1.6 & 0.75 \\
 \hline
 \multicolumn{1}{ c| }{\multirow{2}{*}{\textit{$B$}}} & \multicolumn{1}{ c| }{avg} & 0.68 & 1.47 & 1.24 & 0.33 \\
\multicolumn{1}{ c| }{} & \multicolumn{1}{ c| }{md} & 0.18 & 0 & 0.56 & 0 \\
 \hline
 \multicolumn{1}{ c| }{\multirow{2}{*}{\textit{$Id$}}} & \multicolumn{1}{ c| }{avg} & 0.35 & 1.74 & 0.75 & 0.52 \\
\multicolumn{1}{ c| }{} & \multicolumn{1}{ c| }{md} & 0 & 1.131 & 1.67 & 0 \\
 \hline
\end{tabular}
}
\end{table}

Figure~\ref{fig:landmarkSkull} presents three experimental results when there is high morphological variation and deformity. In Figure~\ref{fig:abnormal1}, due to the genioplasty with chin advancement and rigid fixation, there is a protuberance on the mandible distoring the normal anatomy. In Figure~\ref{fig:abnormal2}, condyle-ramus unit is absent  on the left side of the mandible due to a congenital birth defect. The Geodesic Landmark Localization network successfully detected $4$ landmarks. Note that the fifth landmark was on the missing bone, and it was not located as an outcome of the landmarking process. This is one of the strengths of the proposed method. In Figure~\ref{fig:abnormal3}, the patient had bilateral surgical implants along the ascending ramus (bicortical positional screws), and bilateral condyle and coronoid processes are fixed with these implants. The landmarking process was successful even in this challenging case.

\begin{figure}[H]
\centering
\includegraphics[width=1\linewidth]{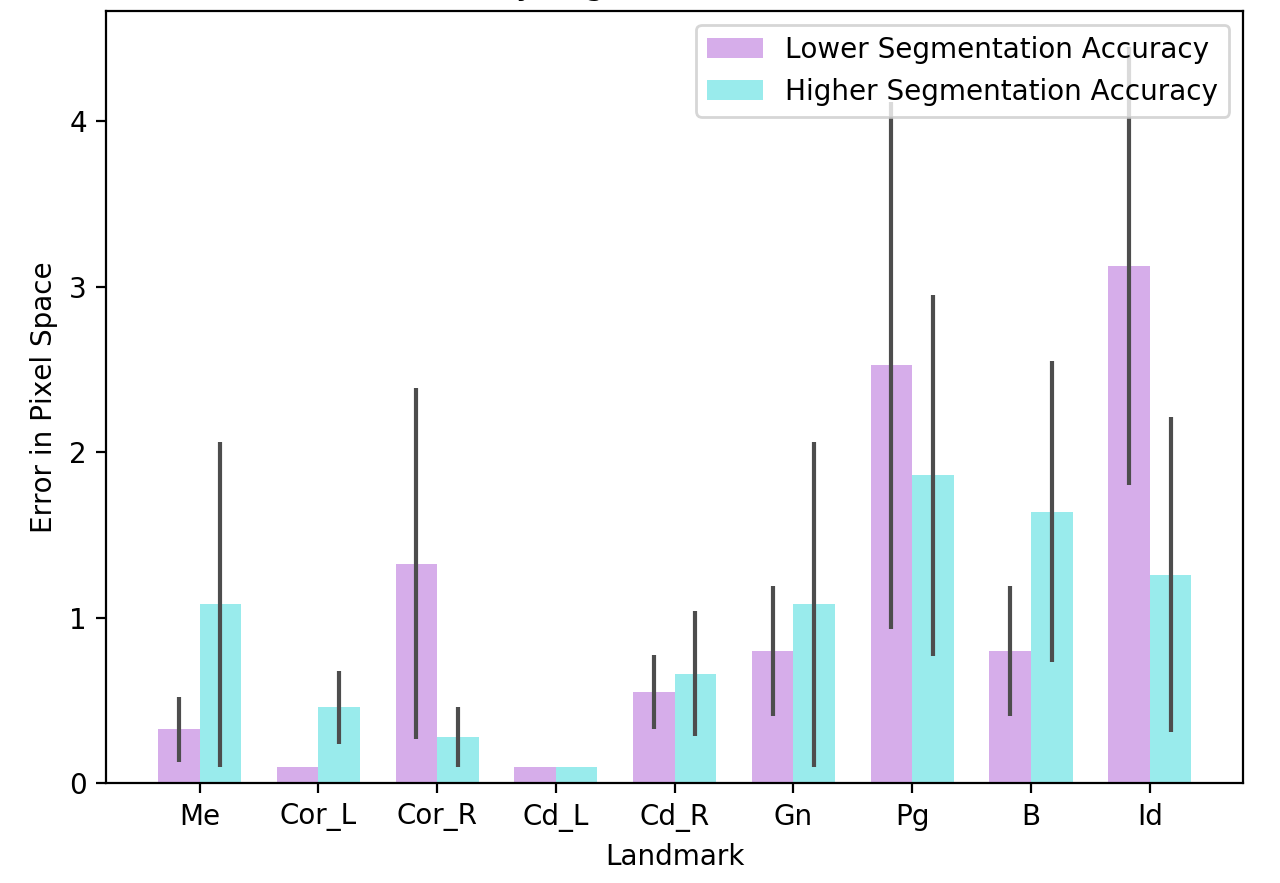}
\caption{Impact of segmentation accuracy  on the landmark localization process.}
\label{fig:segImpactLandmarkAcc}
\end{figure}

We also evaluated the impact of segmentation accuracy on the landmark localization error (Figure~\ref{fig:segImpactLandmarkAcc}). In this evaluation, we first grouped the testing scans into $2$ groups according to their dice values as lower and higher segmentation accuracies (i.e., $\leq 90\%$ as lower, $>90\%$ as higher). Next, we compared the landmark localization errors in pixel space for these two groups. In Figure~\ref{fig:segImpactLandmarkAcc}, the landmarking process was robust to changes in segmentation accuracy, and never achieves greater than 3 pixels errors. It should be also noted that the mean and median segmentation accuracy were still very high in our experiments, leading to successful landmark localizations even at the low end of the dice values. 
Unlike this high robustness, we also noted that the errors in the $2^{nd}$ module (sparsely-spaced landmark localization module) can be propagated to the $3^{rd}$ module (closely-spaced landmark localization module). That is, on a scan with sagittal slice spacing \textbf{d}, if ``Menton'' resides on the sagittal slice \textbf{s} and localized by the $2^{nd}$ module at the sagittal slice \textbf{s-n}, then the error  $\mathbf{(n \times d)}$ mm is propagated to all closely-spaced landmarks. Overall, the landmark localization is robust to the segmentation step and a potential (visible) error can happen only when the Menton (closely-spaced landmark) is located incorrectly due to a potential segmentation error.

Table~\ref{table:regularization} summarizes the average and median errors of localized landmarks in millimeters with respect to different regularization methods. Since the landmarking process is a detection problem, we anticipated using different pooling strategies to evaluate the network performance. We observed that max pooling consistently outperformed other regularization methods. Unlike the segmentation problem, where average pooling was most effective in pixel level predictions, landmarking was driven by discriminative features, enhanced by max pool operation. All average and median errors of the landmark localizations were within the clinically acceptable limits (less than 3mm, i.e., submillimetric errors). 

\section{Discussion and Conclusion}

Overall, the proposed networks (Tiramisu and improved Zhang's U-Net) have enjoyed fast convergence (around 20 epochs) and high accuracy in a very challenging CBCT dataset. Tiramisu was observed to have better converging and training ability compared to improved Zhang's U-Net. For landmark localization, improved Zhang's U-Net in the geodesic space has performed comparably to the validated operator manual landmarking (e.g., median displacement error of 0 mm in most landmarks). 

Fully convolution network (FCN)~\cite{fcn} has significantly changed the landscape of semantic image segmentation frameworks. Based on the FCN, Ronneberger et al.~\cite{unet2015} introduced the U-Net which became the baseline for the current medical image segmentation tasks. The literature for particular medical image segmentation applications based on U-Net is vast; employing the encoder-decoder structure, dense connections, skip connections, residual blocks, and other types of architectural additions to improve segmentation accuracies for particular medical imaging applications. One major drawback of the U-Net framework is the inefficiency introduced by the significantly higher number of parameters to learn~\cite{rodney_capsules}. Hence, there is an anticipation for improvements in the efficiency and robustness of the U-Net type of architecture in the medical imaging field in the near future. One example of such studies, called Capsules~\cite{rodney_capsules}, may be a good future alternative to what we propose herein.

It is worth noting that authors in~\cite{zhang_2017} have considered the landmarking and segmentation processes in the same platform as a multi-task learning problem. In our study, we have focused on individual aspects of segmentation and landmarking, and have proposed novel architectural designs to address problems in both processes that have not been corrected in currently available systems. The natural extension of our work will be to formulate segmentation and landmarking problem within the multi-task learning algorithm, similar to the one proposed by Zhang et al.~\cite{zhang_2017}. 

There are some limitations to our proposed method that should be discussed. Similar to the state-of-the-art method described by Ronnebeger et al. in~\cite{unet2015}, due to memory and hardware limitations, the problems setforth in our study were solved in pseudo-3D. A possible extension of our study will be to work on completely 3D space once hardware and memory supports are available. Another limitation of our work is utilizing a two-cascaded system for landmark localization. We anticipate that locating sparsely and closely-spaced landmarks can potentially be solved in a single network as well. In our method, we observed that when landmarks were very close to each other, combining geodesic distance maps for finding positions of all landmarks became a challenging problem. The hard-coded minimum function that we used for combining geodesic distances created additional artificial landmarks between those closely distributed landmarks. To overcome this problem, we showed a practical and novel use of LSTM-based algorithm to learn the locations of closely-spaced landmarks and avoided such problems. Exploration of different functions other than hard-coded minimum for closely-spaced landmark localization is  subject to further theoretical investigation in geodesic distance maps. 


To further test the algorithms, future studies will include utilization for large cohort landmarking and analysis to establish normative craniofacial datasets.  This fully automated method will enhance high throughput analysis of large, population-based cohorts.  Additionally, studies on rare craniofacial disorders that often have anatomical variation will greatly benefit from the highly accurate landmark localization process.

In summary, our findings suggest that learning-based segmentation and landmarking is a powerful tool for clinical application where anatomical variability is challenging, as in the case of CMF deformity analysis. By carefully designing novel deep learning architectures, we investigated both segmentation and landmarking process in-depth, and presented highly accurate and efficient results derived from 50 extremely challenging CBCT scans. We also addressed some of the poorly understood concepts in deep network architecture (particularly designed for medical image analysis applications) such as the use of dropout and pooling functions for regularization, activation functions for modeling non-linearity, and growth rate for information flow in densely connected layers (See Appendix~\ref{sec:parameters}).


\begin{table*}[t]
\captionsetup{labelfont={color=black},font={color=black}}
\caption{Comparison of segmentation accuracies with respect to different regularization choices. Drop ratio of 0 denotes ``no'' use of dropout layer.}
\label{table:segReg1}
\noindent\resizebox{2.4\columnwidth}{!}{
	\begin{tabular*}{\textwidth}{lccccccccc}
    \cline{1-10}
 	\multicolumn{2}{ c| }{\textbf{Pooling}}&max pool & max pool & avg pool & avg pool & stoc. pool & stoc. pool & avg pool & avg pool\\
        \multicolumn{2}{ c| }{\textbf{Activation}}&  ReLU& ReLU & ReLU & ReLU&  ReLU& ReLU & SWISH & SWISH \\
        \multicolumn{2}{ c| }{\textbf{Growth Rate}} & 16 & 24& 16 & 24 & 16 & 24 & 16 & 24\\
        \cline{1-10}
 	\multicolumn{1}{ c| }{\multirow{2}{*}{\textit{$\textbf{DSC(\%)}$}}}	& \multicolumn{1}{ c| }{drop ratio=0.0} & 93.09 & 92.64 & 93.16 & 92.16 & 92.59 & 90.93 &  93.14 & 92.60\\
    \cline{2-10}
        \multicolumn{1}{ c| }{\multirow{2}{*}{}}	& \multicolumn{1}{ c| }{drop ratio=0.2} & 93.10 & 93.08 & 93.36 & \textbf{93.82} & 92.14 & 92.53 & 91.79 & 93.67\\
        \cline{1-10}
	\end{tabular*}
}
\end{table*}


\begin{table}[ht]
\captionsetup{labelfont={color=black},font={color=black}}
\centering
\caption{Resulting segmentation DSC accuracies with respect to the drop ratio (avg pooling, ReLU, and growth rate of 24). Note that drop ratio of 0 denotes ``no'' use of dropout layer.}
\label{table:segReg}
\resizebox{\columnwidth}{!}{
	\begin{tabular}{cccccc}
		\hline
        \textbf{Drop Ratio} & 0.5 & 0.3 & 0.2 & 0.1 & 0.0  \\
 		\hline
 		 \textbf{DSC(\%)} & 91.21 & 93.37 &  \textbf{93.82} & 92.90 & 92.88 \\
 		\hline
	\end{tabular}
}
\end{table}

\begin{table}[h]
\captionsetup{labelfont={color=black},font={color=black}}
\centering
\caption{Effect of different growth rates on segmentation performance using Tiramisu with avg. pooling.}
\label{table:segGrowthRate}
\resizebox{\columnwidth}{!}{
	\begin{tabular}{ccccc}
		\hline
        \textbf{Growth Rate ($k$)} & 12 &  16 & 24 & 32 \\ 
 		\hline
 		\textbf{DSC(\%)} & 92.63 &  93.36 &  \textbf{93.82} & 92.60 \\ 
        \textbf{HD(mm)}  & 6.44 & 5.50 & 5.47 & \textbf{5.02} \\ 
 		\hline
	\end{tabular}
}
\end{table}

\appendix
\subsection{Evaluation of the Segmentation Network Parameters}
\label{sec:parameters}

\subsubsection{Effect of pooling functions}
After extensive experimental comparisons, we found that average pooling acts as a robust regularizer compared to other pooling functions such as max pooling and stochastic pooling (Table~\ref{table:segReg}). 
\subsubsection{Disharmony between BN and dropout}
We found that when BN is used in the network for segmentation purpose, the use of dropout is often detrimental except for only a drop rate of 20\%. Similarly, we found that average pooling was the most robust pooling function compared to others when BN and dropout were used together. 
\subsubsection{The role of growth rate in dense blocks}
Tiramisu network with 103 layers (growth rate of 16) has a proven success in the computer vision tasks. However, in our experiments, we observed that a Tiramisu network with a growth rate of 24 and drop rate of 0.2 produces the best accuracies (See Table~\ref{table:segGrowthRate}). Further, when no dropout is used (drop rate is 0), the growth rate performance inverses (See Table~\ref{table:segReg1}), implying the regularizing impact of employing dropout on the neural networks.
\subsubsection{The choice of activation functions} 
Although there have been many hand-designed activation functions proposed for deep networks, ReLU (rectified linear unit) became an almost standard choice for most CNNs. The main reason is due to its significant effect on the training dynamics and high task performances. More recently, another activation function, called ``Swish"~\cite{ramachandran2018searching}, was proposed. Unlike other activation functions, Swish was automatically determined based on a combination of exhaustive and reinforcement learning-based search. Authors showed that Swish tend to perform better than ReLU for very deep models. Since the proposed Tiramisu has 103 layers, we replaced all ReLU functions with Swish, which is a weighted sigmoid function $f(x)=x.sigmoid(\beta x)$, and explored the network behaviors. We summarized the network performance in Table~\ref{table:segReg1}. Overall, we did not observe significant differences between ReLU and Swish, but ReLU led into slightly better results in all sub-experiments.

\subsection{Qualitative Evaluation}
\begin{figure}[t]
\centering
\includegraphics[width=0.8\linewidth,valign=t]{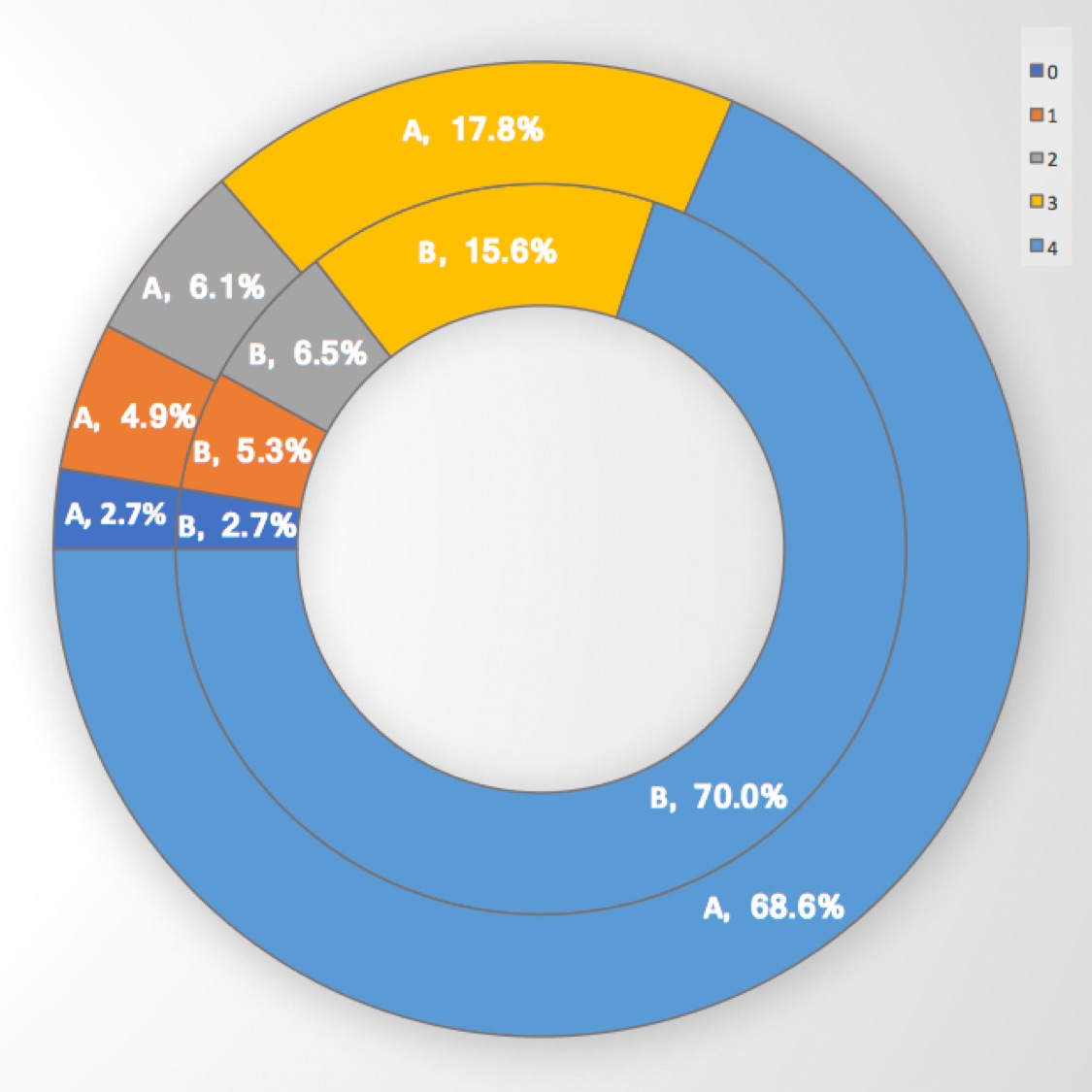}
\caption{Qualitative evaluation of $250$ scans from the NIDCR/NIH dataset by $2$ experts, \textbf{A} and \textbf{B}}.
\label{fig:visual_perm}
\end{figure}

\begin {figure*}[]
\captionsetup{labelfont={color=black},font={color=black}}
\centering
  \begin{tabular}{ccc}
  \hline
  \hline \\[0.25ex]
  \multicolumn{3}{c}{(a) Score 4}\\[1ex]
    \includegraphics[width=0.22\linewidth]{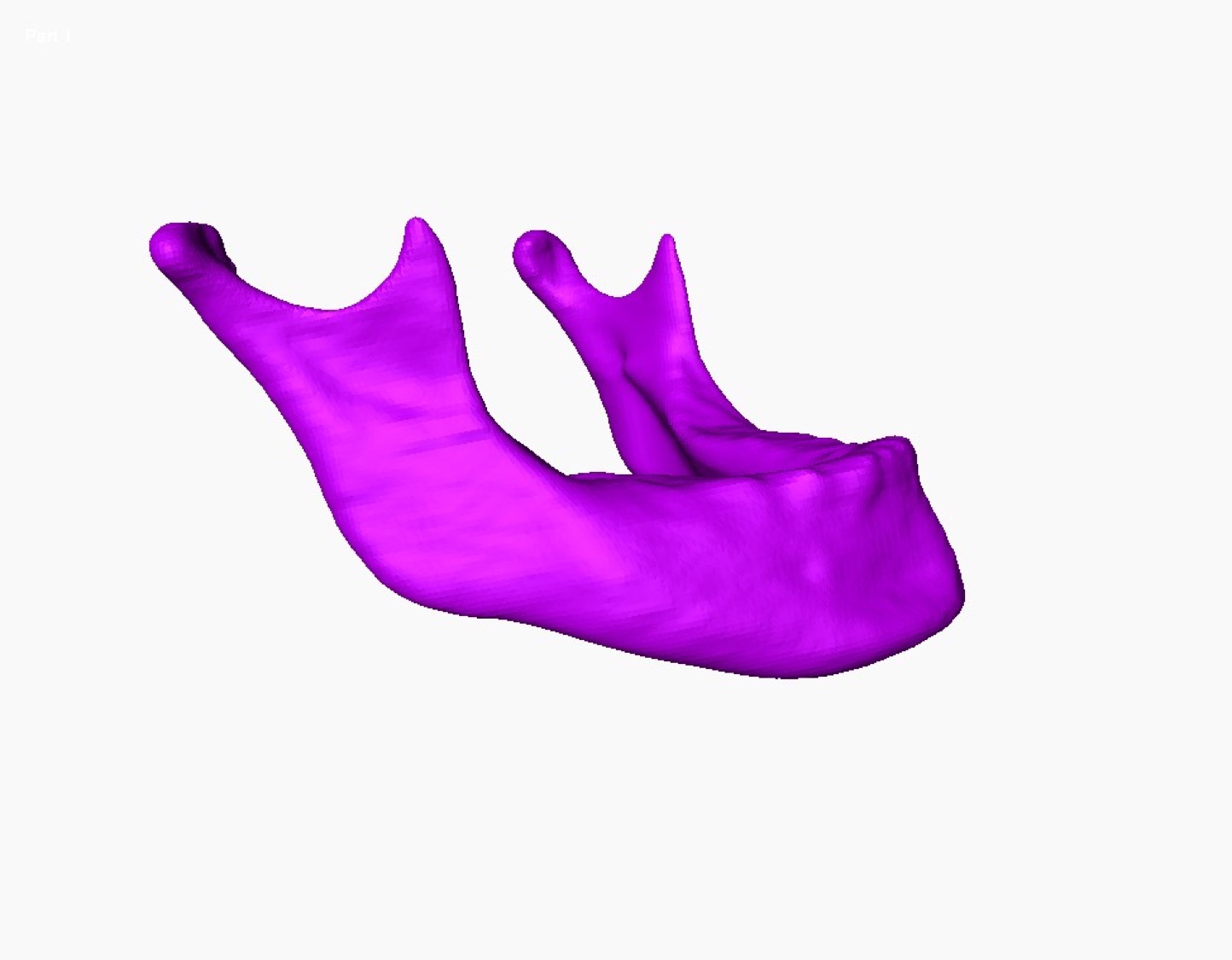} & 		      \includegraphics[width=0.22\linewidth]{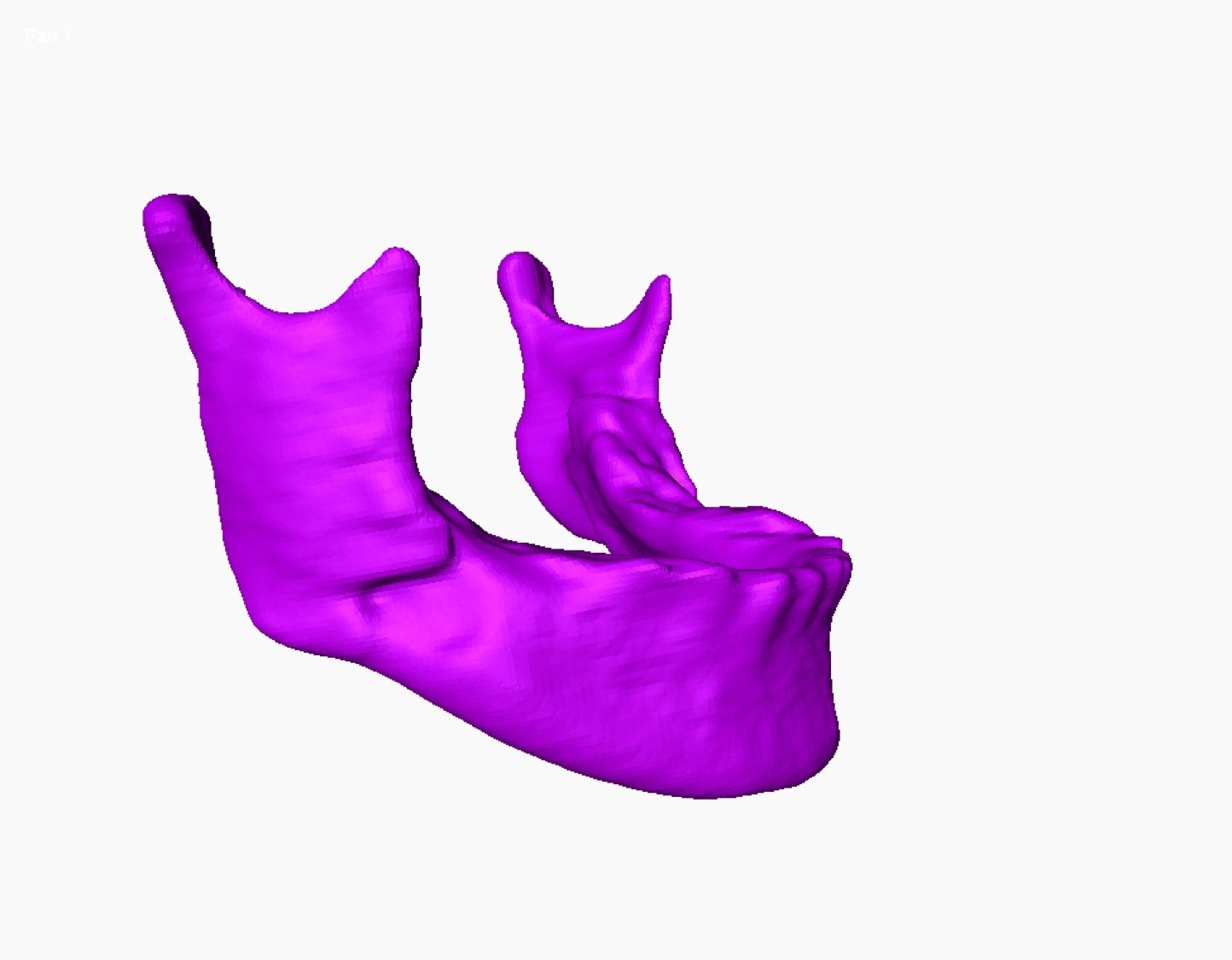}  &
    \includegraphics[width=0.22\linewidth]{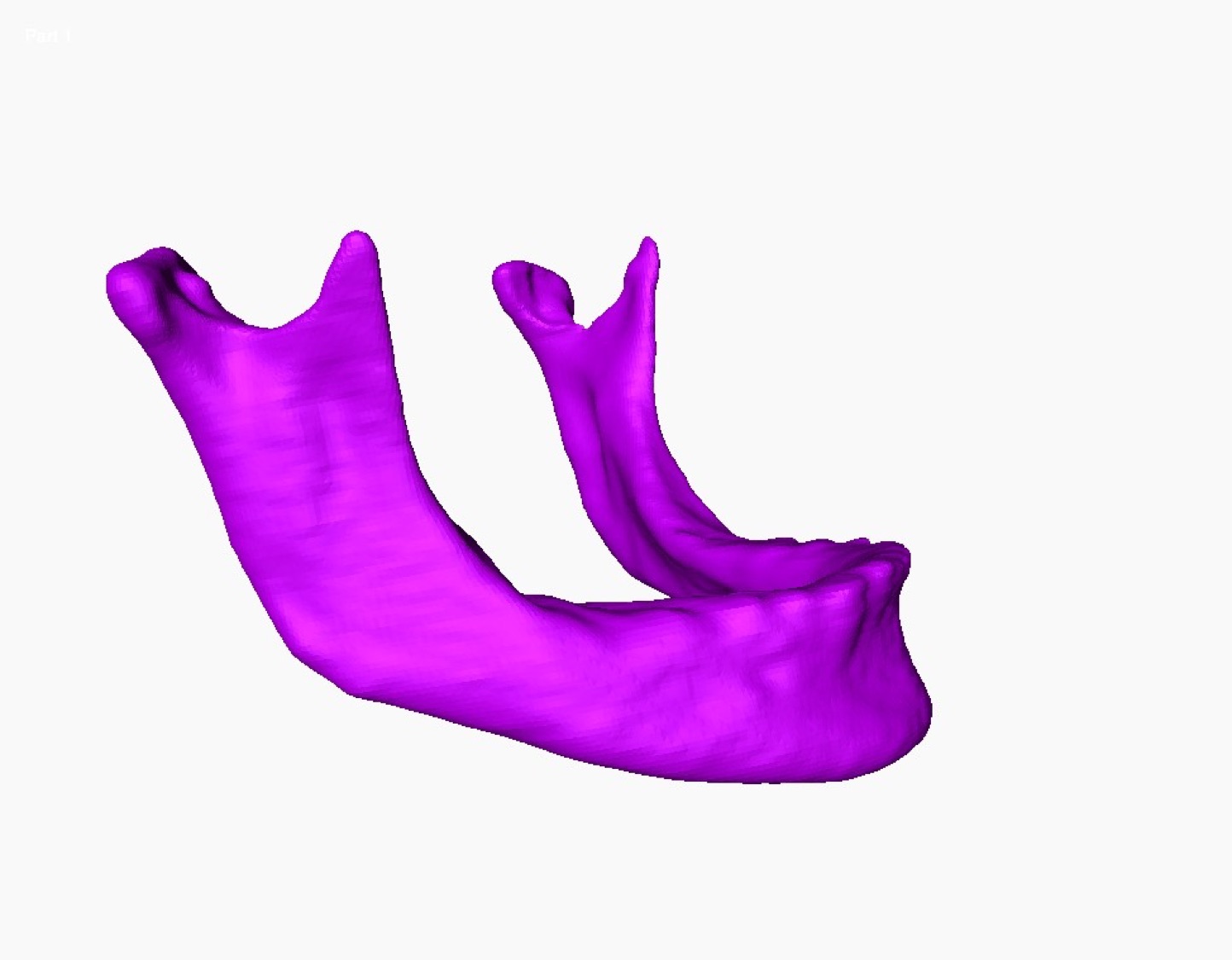} \\
    (a1) & (a2) & (a3) \\[1ex]
    \includegraphics[width=0.22\linewidth]{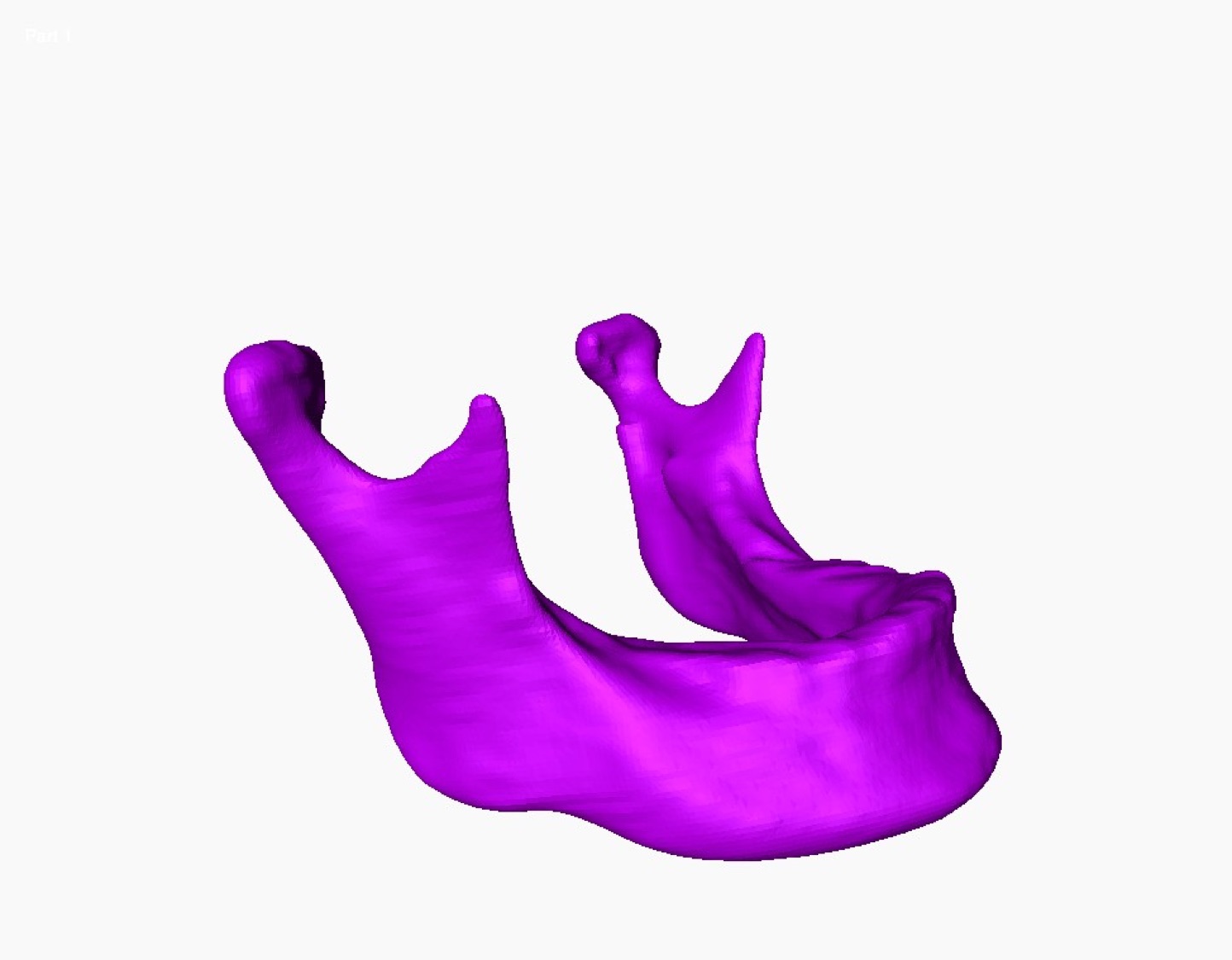} & \includegraphics[width=0.22\linewidth]{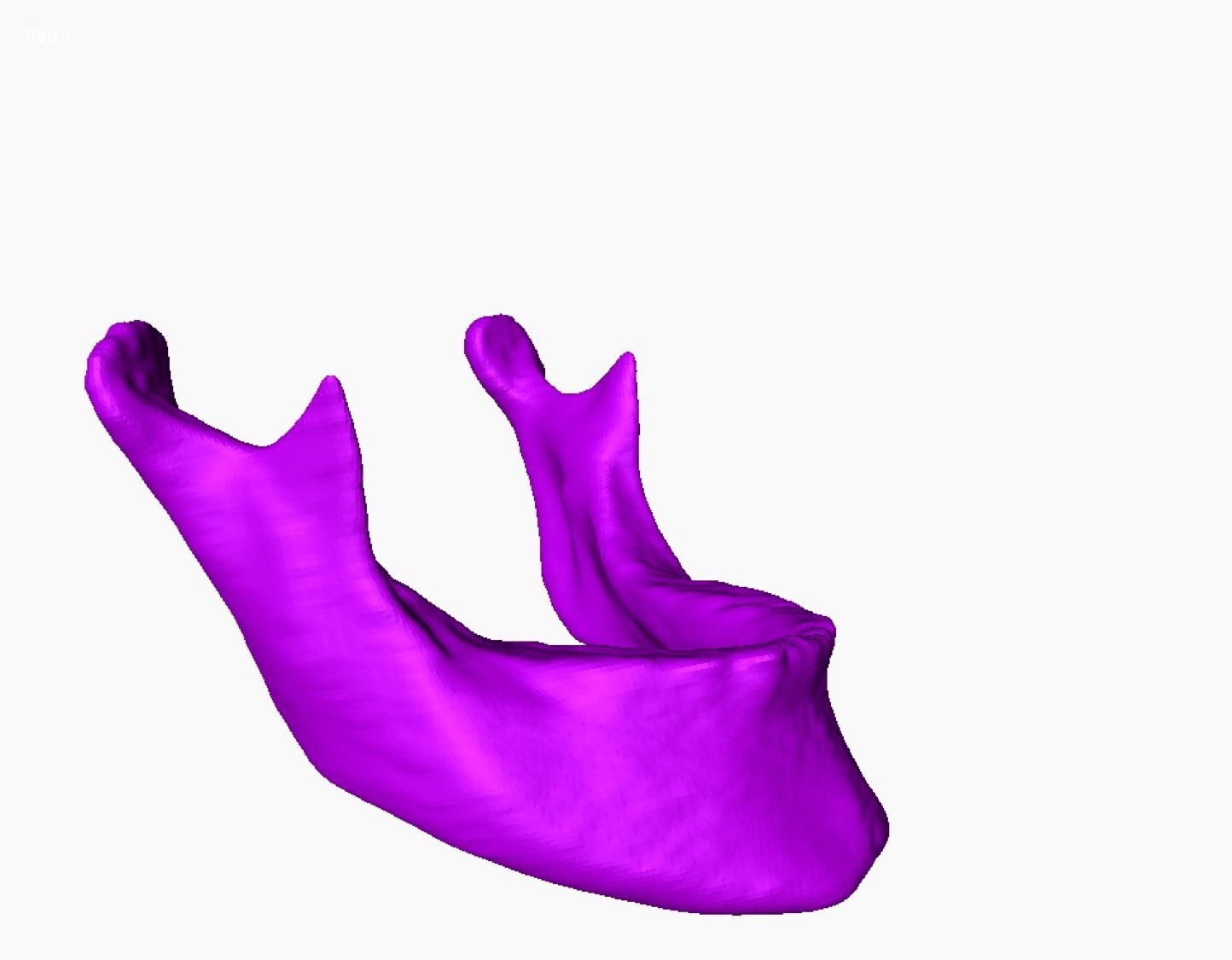} &
    \includegraphics[width=0.22\linewidth]{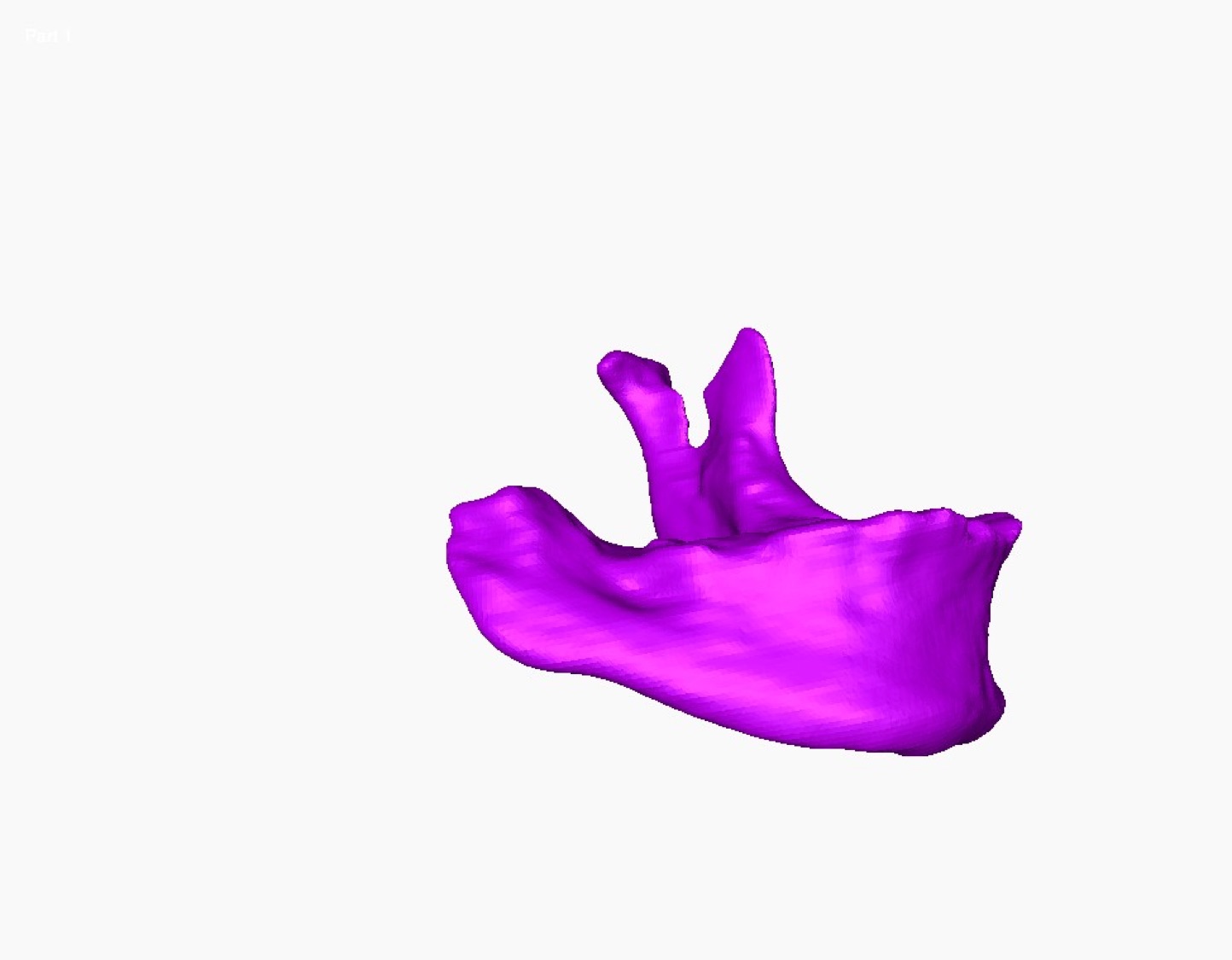} \\
    (a4) & (a5) & (a6) \\[1ex] 
      \hline \\[0.25ex]
      \multicolumn{3}{c}{(b) Score 3}\\[1ex]
    \includegraphics[width=0.22\linewidth]{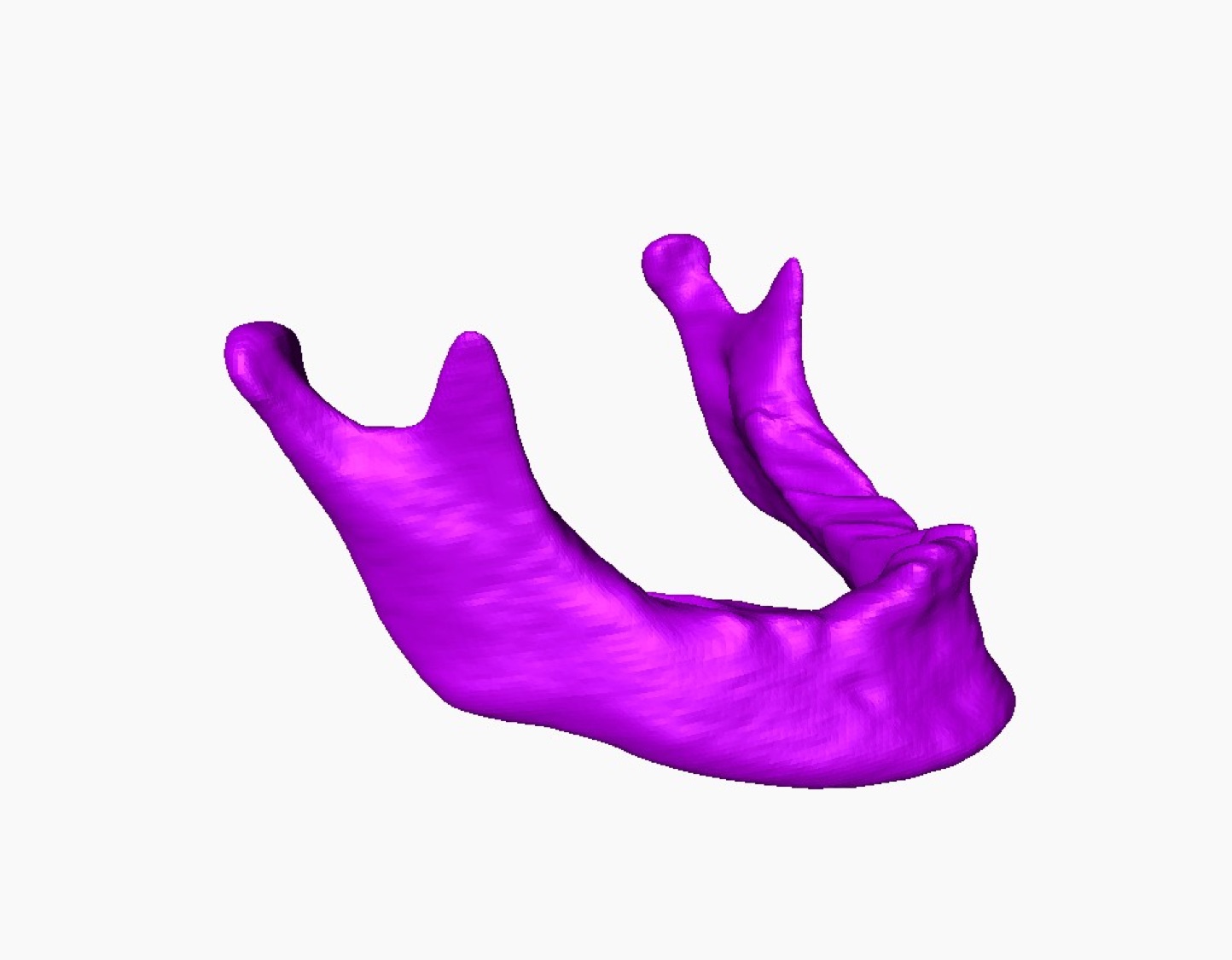} & 		      \includegraphics[width=0.22\linewidth]{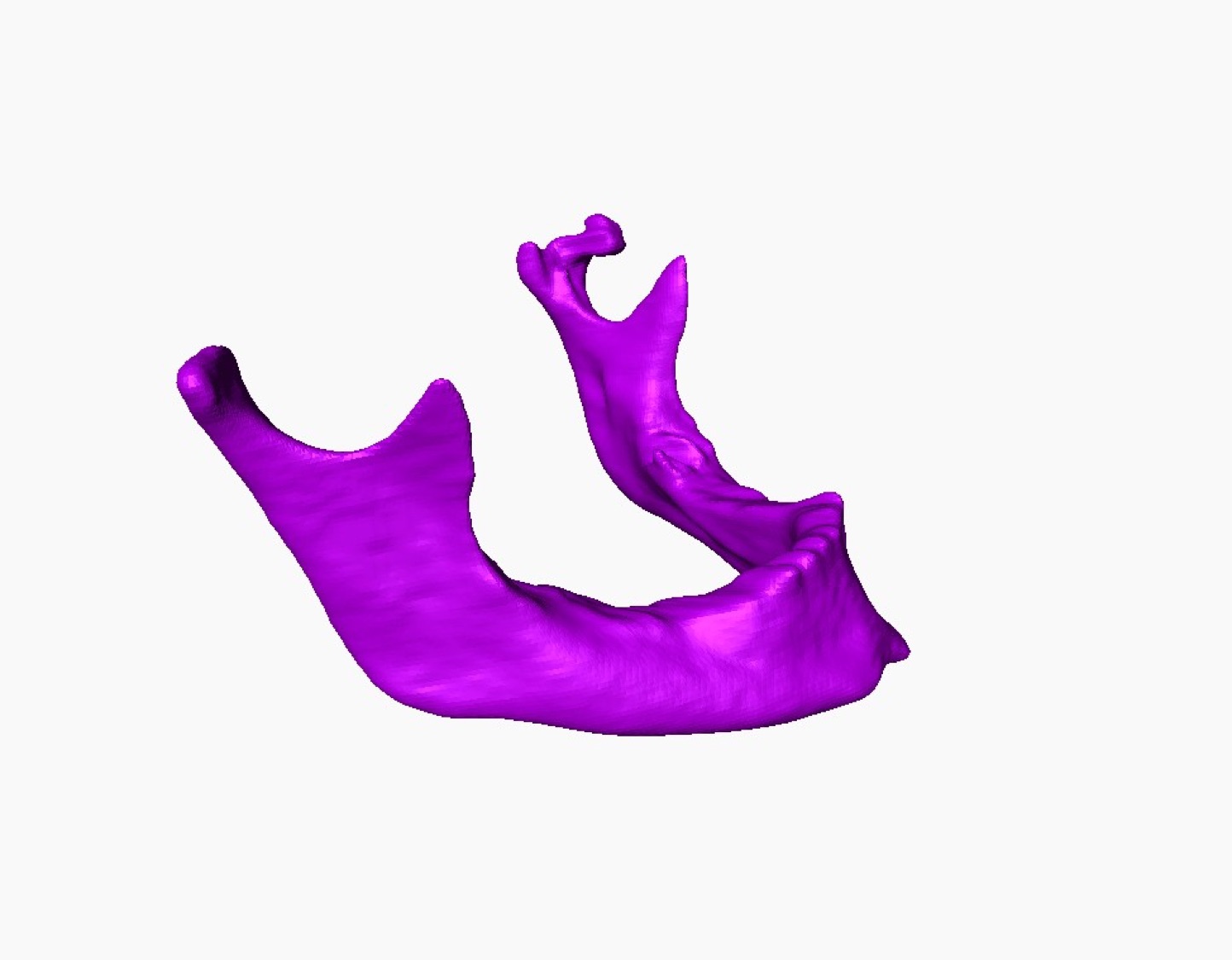}  &
    \includegraphics[width=0.22\linewidth]{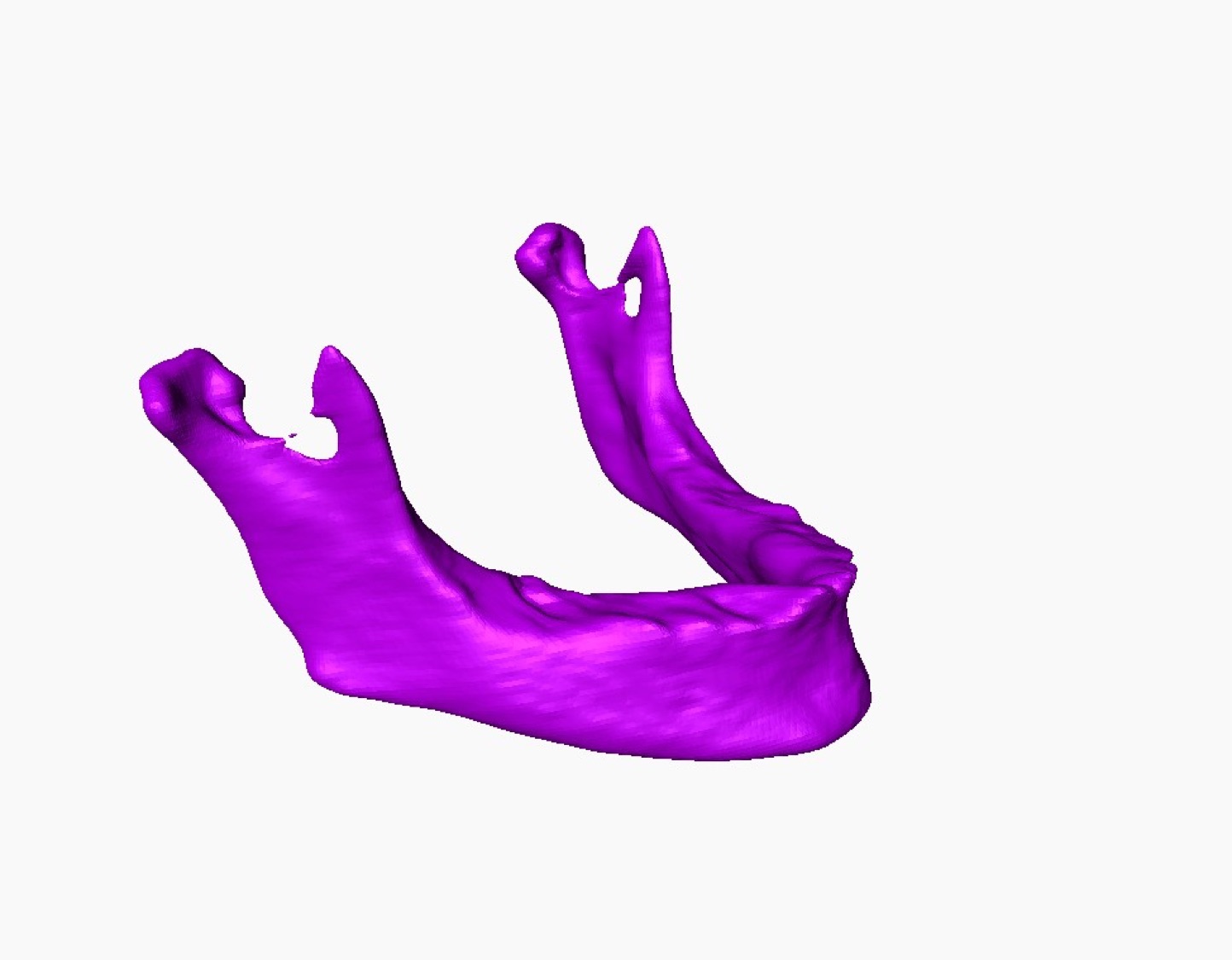} \\
    (b1) & (b2) & (b3) \\[1ex]
      \hline \\[0.25ex]
      \multicolumn{3}{c}{(c) Score 2}\\[1ex]
          \includegraphics[width=0.22\linewidth]{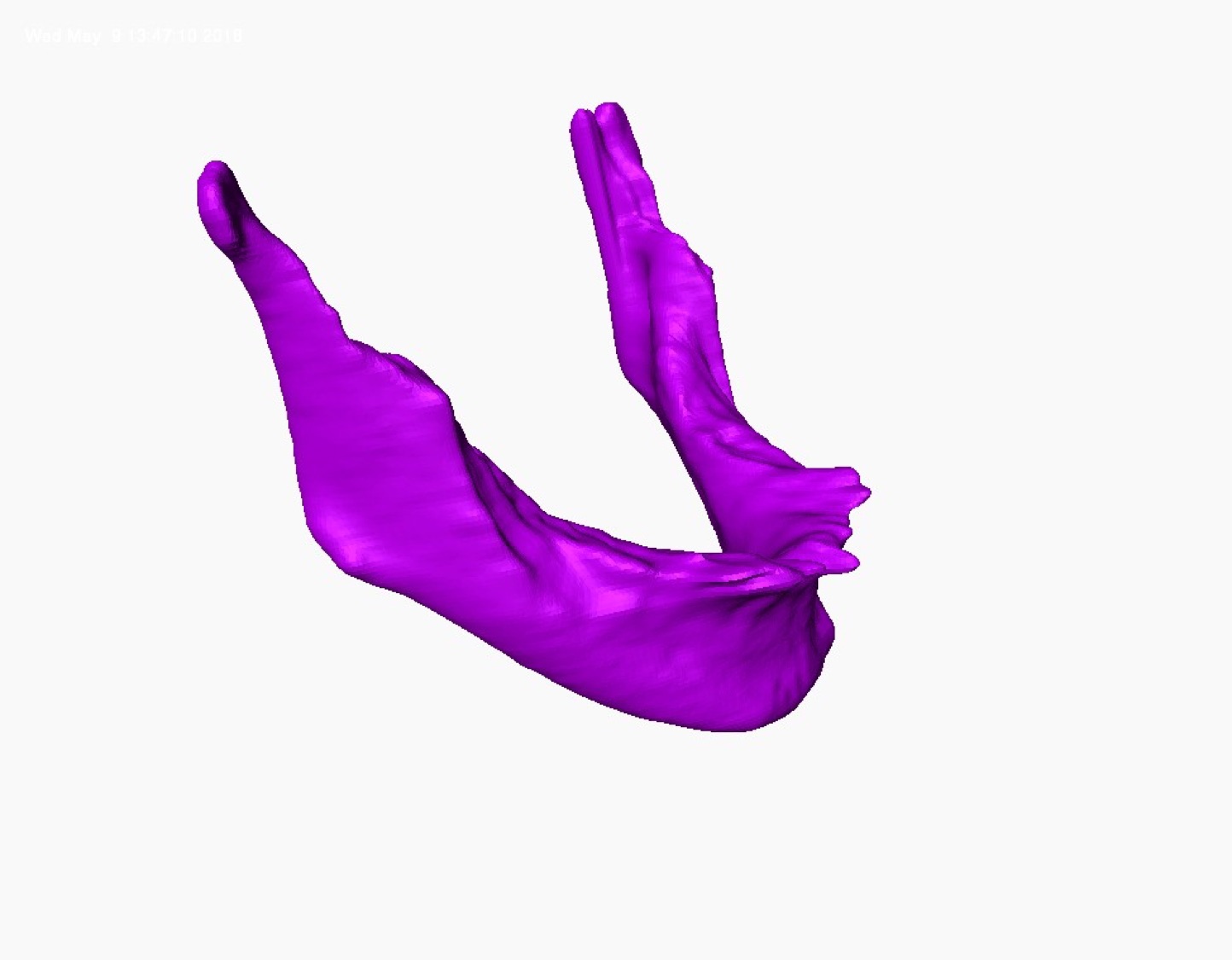} & 		  \includegraphics[width=0.22\linewidth]{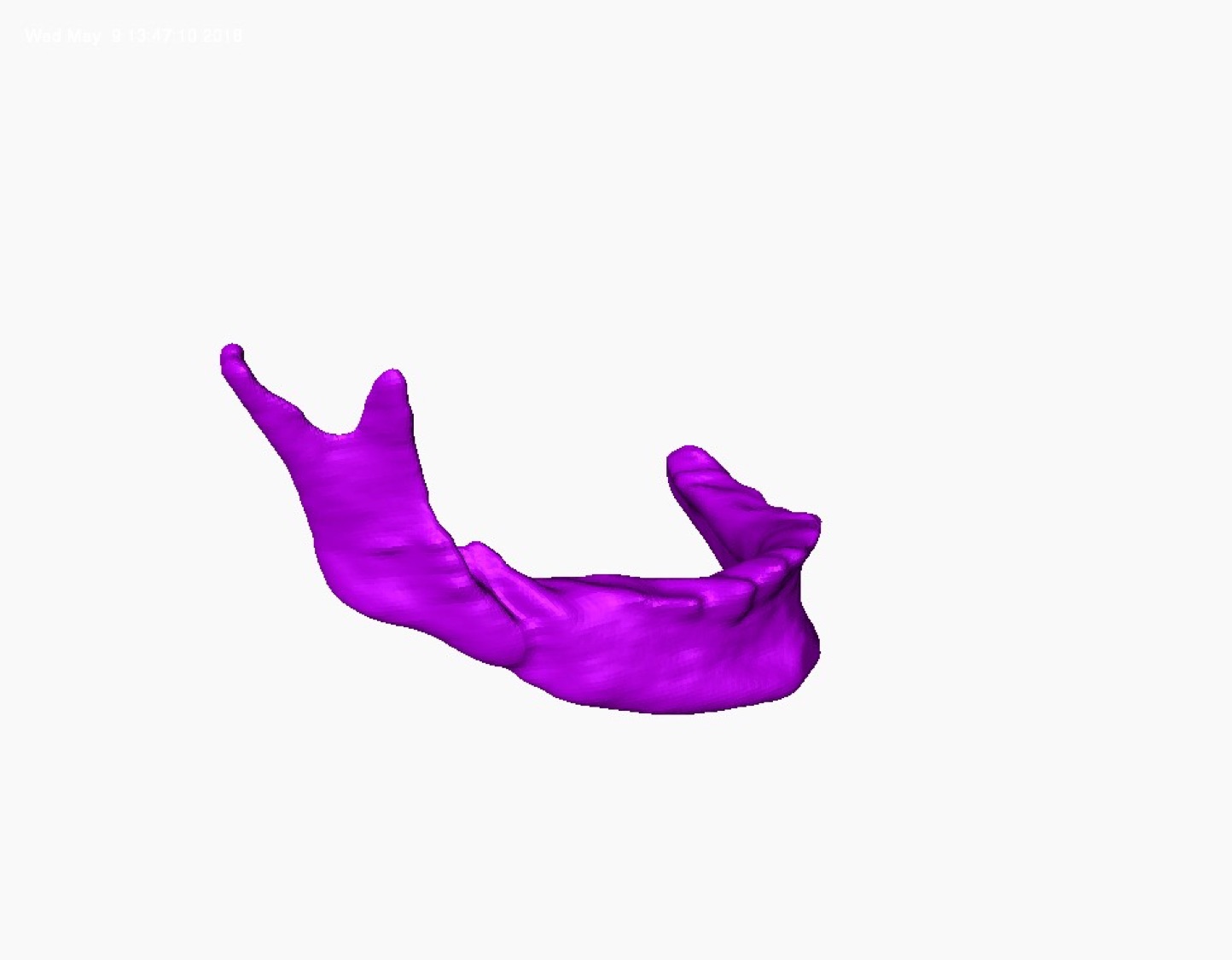}  &
    \includegraphics[width=0.22\linewidth]{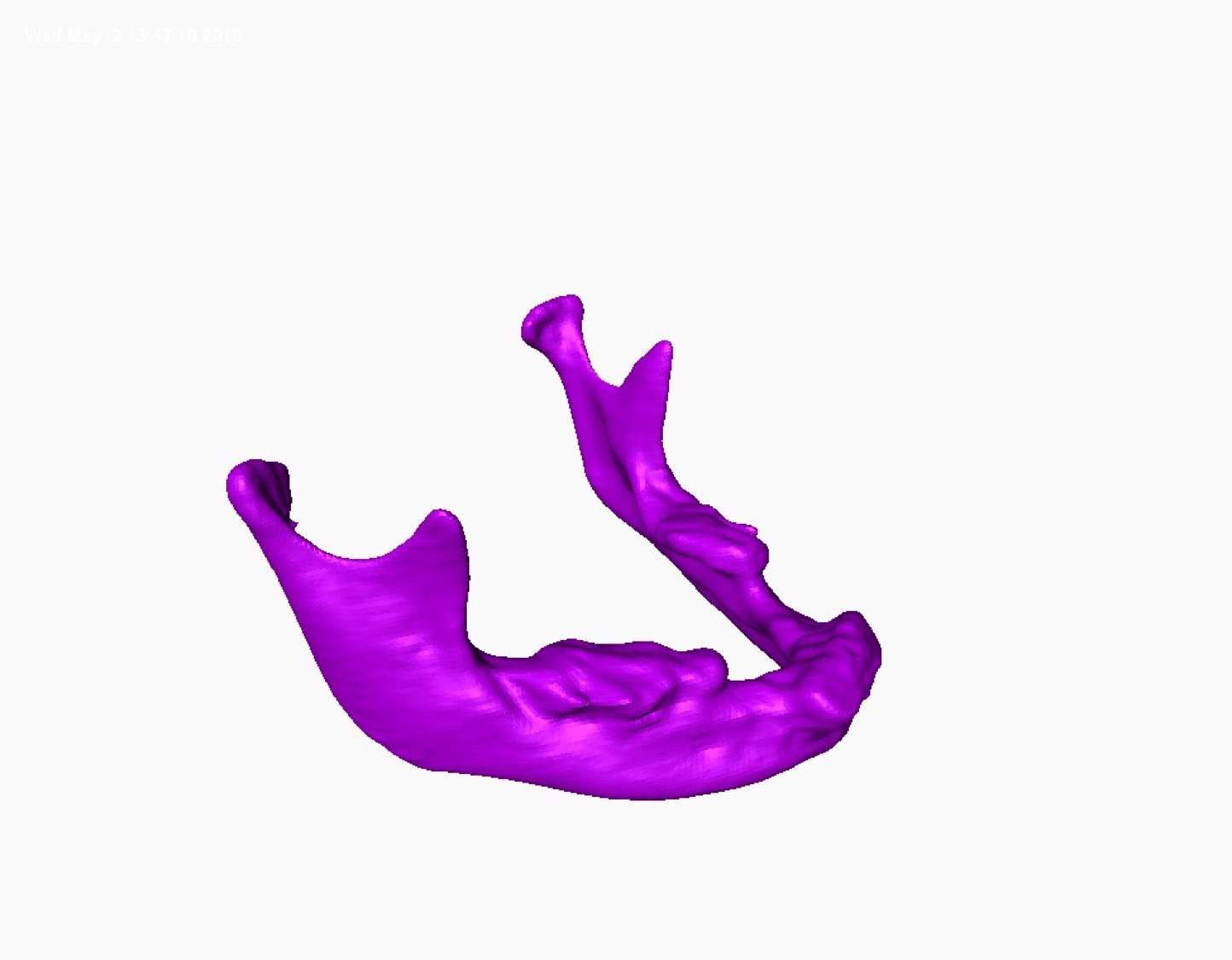} \\
     (c1) & (c2) & (c3) \\[1ex]
      \hline \\[0.25ex]
       \multicolumn{3}{c}{(d) Score 1}\\[1ex]
          \includegraphics[width=0.22\linewidth]{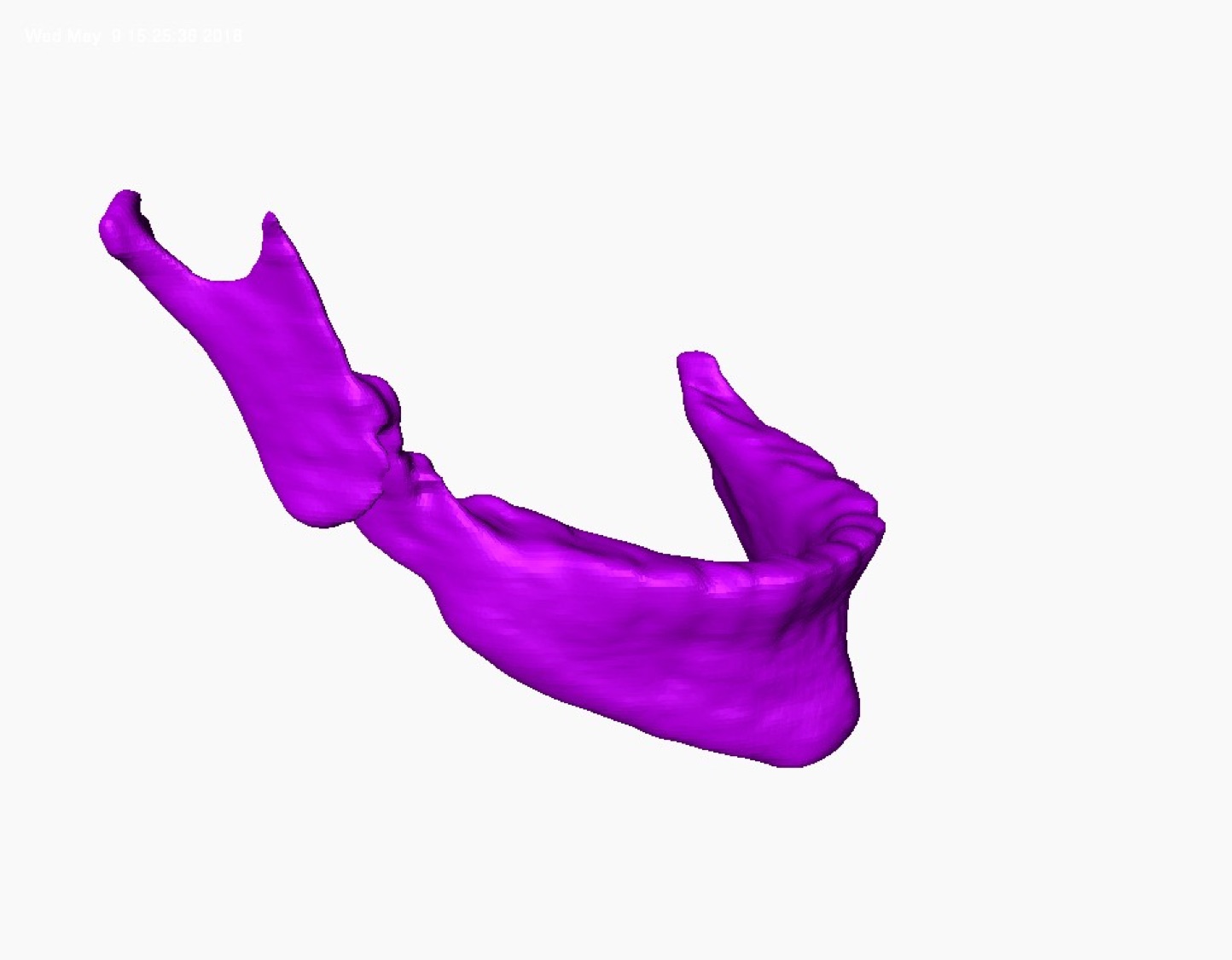} & 		  \includegraphics[width=0.22\linewidth]{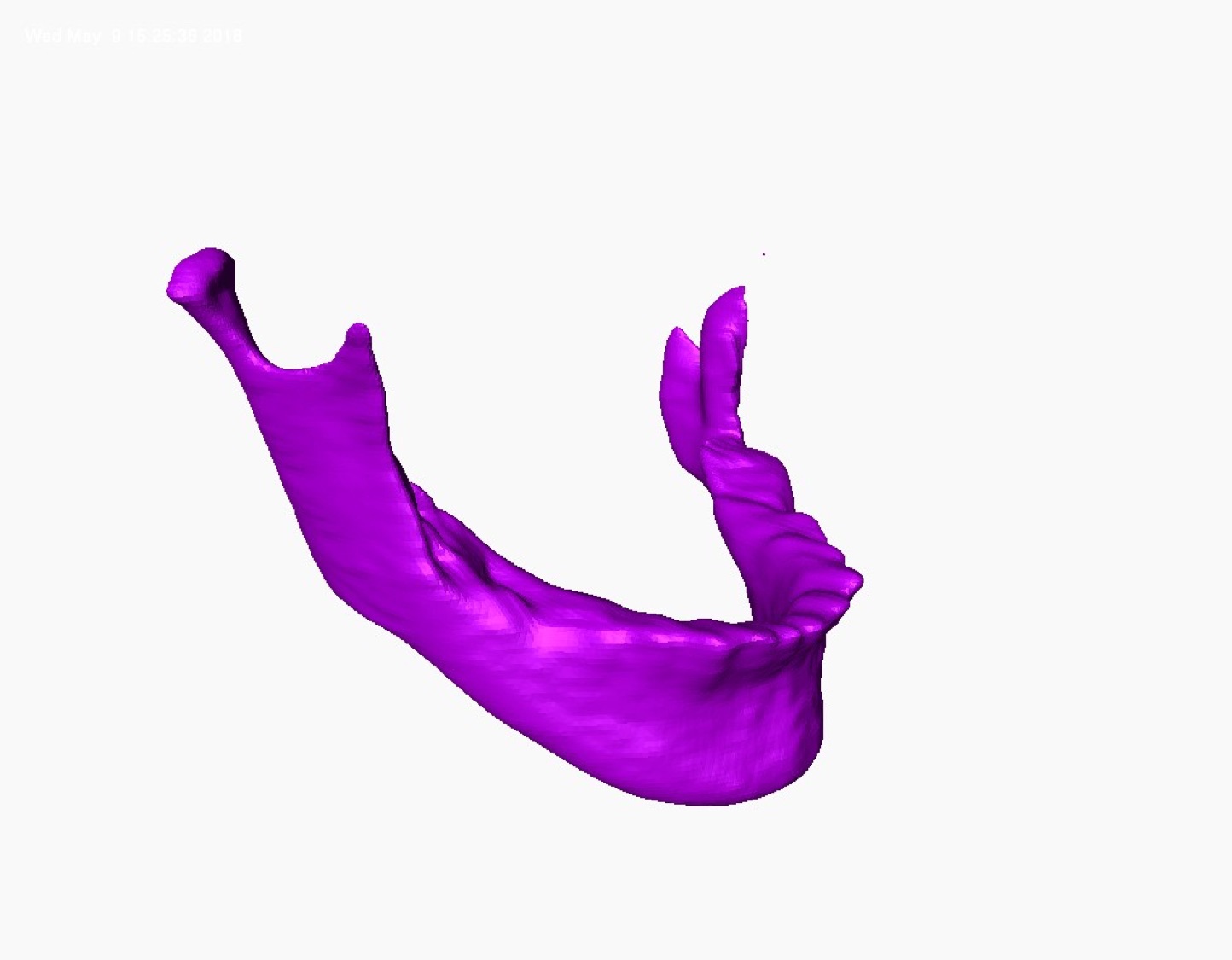}  &
    \includegraphics[width=0.22\linewidth]{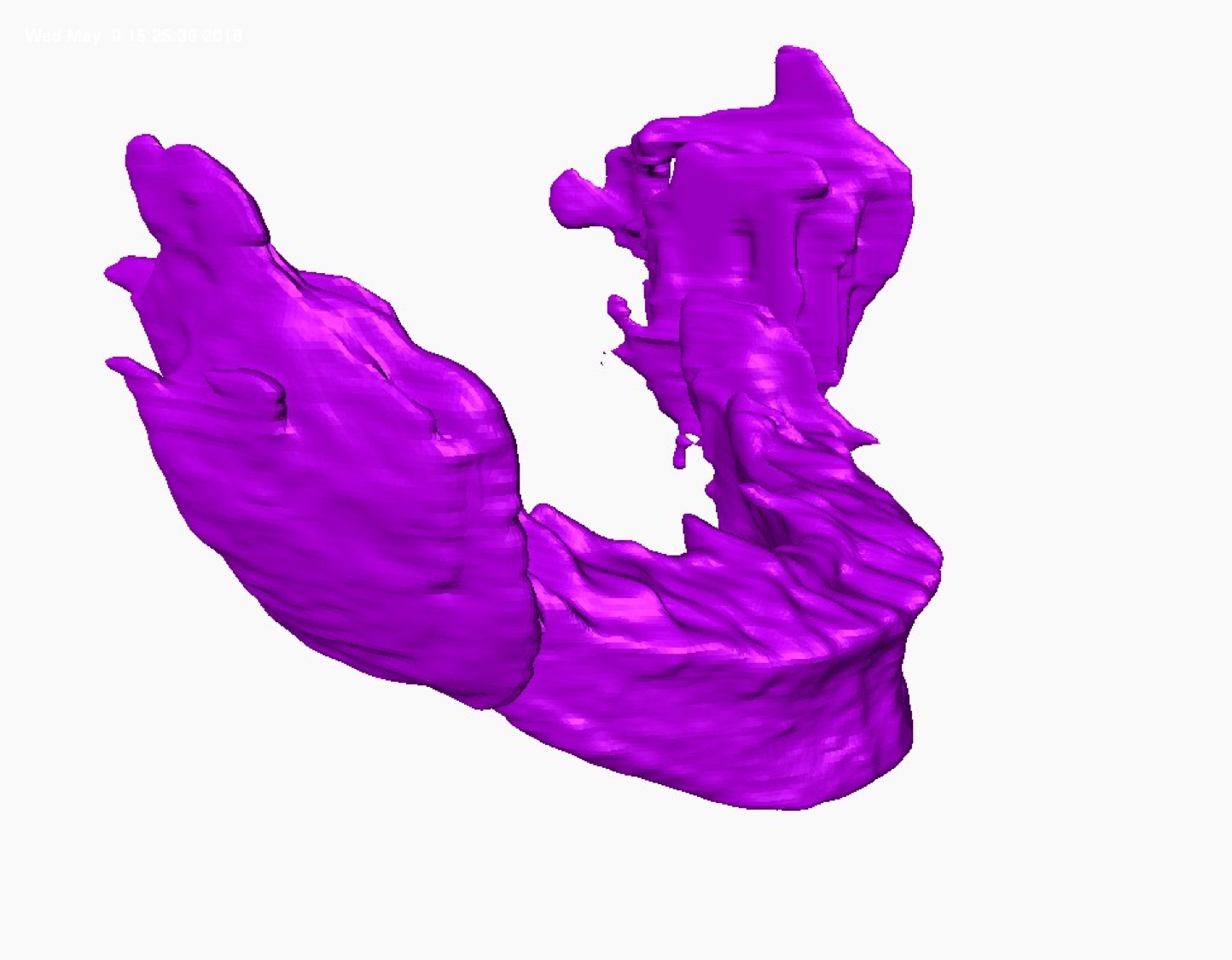} \\
     (d1) & (d2) & (d3) \\[1ex]
    \hline
  \hline \\[0.25ex]
\end{tabular}
\caption{Qualitative Evaluation Scores of the Segmentation Results. The experts visually evaluated the performance of the segmentation of the $250$ patient scans in the score range $1$ to $4$, where $1$ is inferior. Examples of scans with scores $1$-$4$ are presented.}
\label{fig:visualEvalScores}
\end{figure*}

The availability of head-neck CBCT scans with manual annotations is very limited. Our total CBCT dataset is composed of $250$ patient CBCT images provided by our collaborators at the NIDCR/NIH. However, only 50 of them were manually annotated by the three experts. Hence, to measure the performance of the algorithm on all available scans, by following the routine radiologic evaluation of the scans, two experts visually scored the performance of the segmentations in the range from 0 to 4, where 1 is unacceptable, 2 is borderline, 3 is acceptable at clinical level, and 4 is superior (excellent) (Figures~\ref{fig:visual_perm} and~\ref{fig:visualEvalScores}). When the scan is completely distorted or mandible does not exist in its entirety in the scan, it is not possible to automatically segment mandible, hence a score of 0 is given. 

Scores of $3$ and $4$ represent clinically acceptable segmentations, where score $3$ may correspond to minor deformations in the segmentation. The left top part of the Mandible (Figure~\ref{fig:visualEvalScores}-a6) was missing, for instance, but it was still precisely segmented. The mandible (Figure~\ref{fig:visualEvalScores}-c2), for another example, was composed of two separate parts, and the algorithm detected the larger portion of the mandible in regards to the $3D$ connected component analysis. Further analysis showed that the scans that were scored as $1$ or $2$ were typically the ones with serious anatomical deformations. Approximately $5\%$ of the segmentation results were scored as $1$ by both experts \textbf{A} and \textbf{B} (Figure~\ref{fig:visual_perm}).

\ifCLASSOPTIONcaptionsoff
  \newpage
\fi


\bibliographystyle{IEEEtran}
\bibliography{geodesicLandmark}
\end{document}